\documentclass[pre,12pt,preprint,superscriptaddress, amsmath, amssymb, floatfix]{revtex4-1} 

\renewcommand{\thetable}{\arabic{table}}
\renewcommand{\thefigure}{\arabic{figure}}

\usepackage{tabularray}
\usepackage{graphicx}  
\usepackage{graphics}
\usepackage{dcolumn}  
\usepackage{bm}     
\usepackage{amssymb}  
\usepackage{color}
\usepackage{amsmath}
\usepackage{threeparttable}
\usepackage[font=scriptsize]{caption}
\usepackage{url}
\usepackage{makecell}
\usepackage{diagbox}
\usepackage{booktabs}
\usepackage[figuresright]{rotating}
\usepackage{changes}
\usepackage{hyperref} 
\usepackage{subfigure}
\usepackage{tablefootnote}

\definecolor{darkred}{RGB}{139,0,0}
\definecolor{chartreuse}{RGB}{127,255,0}
\definecolor{goldenrod}{RGB}{218,165,32}
\definecolor{gray}{RGB}{127,127,127}

\definecolor{Magenta}{RGB}{255, 0,255}
\definecolor{Orange}{RGB}{255,165, 0}
\definecolor{Gray}{RGB}{127,127,127}

\begin{document}

\title{Multi-feature concatenation and multi-classifier stacking: an interpretable and generalizable machine learning method for MDD discrimination with rsfMRI}
\author{Yunsong Luo}
\affiliation{College of Computer and Information Science, Southwest University, Chongqing, 400715, P. R. China}

\author{Wenyu Chen}
\affiliation{College of Computer and Information Science, Southwest University, Chongqing, 400715, P. R. China}

\author{Ling Zhan}
\affiliation{College of Computer and Information Science, Southwest University, Chongqing, 400715, P. R. China}

\author{Jiang Qiu}
\affiliation{Key Laboratory of Cognition and Personality(SWU), Ministry of Education, Chongqing, 400715, P. R. China}
            
\affiliation{School of Psychology, Southwest University (SWU), Chongqing,400715, P. R. China}
            
\affiliation{Southwest University Branch, Collaborative Innovation Center of Assessment Toward Basic Education Quality at Beijing Normal University,Chongqing, 400715, P. R. China}  
\author{Tao Jia}
\email{tjia@swu.edu.cn}
\affiliation{College of Computer and Information Science, Southwest University, Chongqing, 400715, P. R. China}

\date{\today}

\begin{abstract}
Major depressive disorder is a serious and heterogeneous psychiatric disorder that needs accurate diagnosis. Resting-state functional MRI (rsfMRI), which captures multiple perspectives on brain structure, function, and connectivity, is increasingly applied in the diagnosis and pathological research of mental diseases. Different machine learning algorithms are then developed to exploit the rich information in rsfMRI and discriminate MDD patients from normal controls. Despite recent advances reported, the discrimination accuracy has room for further improvement. The generalizability and interpretability of the method are not sufficiently addressed either. Here, we propose a machine learning method (MFMC) for MDD discrimination by concatenating multiple features and stacking multiple classifiers. MFMC is tested on the REST-meta-MDD data set that contains 2428 subjects collected from 25 different sites. MFMC yields 96.9$\%$ MDD discrimination accuracy, demonstrating a significant improvement over existing methods. In addition, the generalizability of MFMC is validated by the good performance when the training and testing subjects are from independent sites. The use of XGBoost as the meta classifier allows us to probe the decision process of MFMC. We identify 13 feature values related to 9 brain regions including the posterior cingulate gyrus, superior frontal gyrus orbital part, and angular gyrus, which contribute most to the classification and also demonstrate significant differences at the group level. The use of these 13 feature values alone can reach 87\% of MFMC's full performance when taking all feature values. These features may serve as clinically useful diagnostic and prognostic biomarkers for mental disorders in the future. \


\end{abstract}



\maketitle

\section{Introduction}

Major depressive disorder (MDD) is a mental disease with a high prevalence rate, high suicide rate, and high disability rate, which seriously endangers human physical and mental health and brings tremendous social and economic burden \cite{murray2012disability}. An appropriate diagnosis of MDD is a prerequisite for successful medication \cite{gupta2019machine}. The current clinical diagnosis of MDD primarily depends on symptomatic assessments and treatment response. However, the inherent heterogeneity of MDD may lead to misdiagnosis or delayed diagnosis \cite{schnack2016detecting}. Therefore, there is an urgent need to develop effective diagnostic tools using quantitative measures instead of qualitative symptom descriptions, which improve MDD diagnostic accuracy and provide insights into its pathophysiology \cite{qin2022using}.

Neuroimaging, an advanced technology that provides information on the changes in brain anatomical structure and brain nerve activity, is becoming an effective data source for the diagnosis and pathological research of mental diseases \cite{zatorre2012plasticity}. Resting-state functional MRI (rsfMRI) is of particular significance in many practices due to its capability to measure the temporal correlation of spontaneous signal among spatially distributed brain regions, providing multiple perspectives on brain structure, function, and connectivity maps \cite{gao2018machine}. To fully exploit the rich information captured by rsfMRI, machine learning techniques are applied, which are capable of detecting abnormal patterns that remain invisible at the group-level level and identifying individuals with mental disorders from subtle spatial distribution differences of the brain \cite{nouretdinov2011machine, zhang2016discriminative}. There are several machine learning tools proposed that take rsfMRI data as the input and discriminate MDD patients from normal controls (NC) with excellent accuracy \cite{guo2019resting,yan2020quantitative,jin2020region,shimizu2020maximum,wei2013identifying}.

Despite these advances reported, current studies on machine learning algorithms for MDD discrimination still hold several limitations. Unlike many other performance-driven tasks, a new MDD discrimination method is rarely compared with other state-of-the-art methods when proposed. In general, according to reported performance in current research, the MDD discrimination accuracy has room for further improvement. In addition, abnormal patterns of MDD patients at the group level are reported, but their roles in MDD discrimination remain to be explored. This requires the interpretability of the machine learning algorithm that allows its decision-making process to be understood and explained. Finally, existing studies based on small data sets from one single site may face the risk of overfitting. Such methods may not be successfully transferred and validated in rsfMRI data collected from other sites. To apply the method in real clinical practices, the generalizability needs to be carefully checked, which is not sufficiently emphasized in past studies.

In this study, we propose an MDD discriminant method MFMC by concatenating multiple features and stacking multiple classifiers. The model is trained and validated on the REST-meta-MDD data set that consists of samples collected from 25 sites in China, including a total of 1300 MDD patients and 1100 controls. Despite the relatively straightforward architecture of MFMC, it outperforms existing methods with a significant advance in discrimination accuracy. The generalizability of the model is also validated when the training and testing data are composed of samples from independent sites. In addition, the use of XGBoost as the meta classifier allows us to explain the decision with the Shapley Additive Explanations (SHAP) value \cite{lundberg2020local,zhang2020clinically}, which gives feature values of brain regions that contribute most to the MDD discrimination at the individual level. We identify 13 features that are significantly different at the group level, which preserve a strong predictive capability. These features may serve as clinically useful diagnostic and prognostic biomarkers for mental disorders in the future. We make the codes of our model publicly available for future references and comparisons.

\section{Materials and Methods}
\subsection{Subjects}
REST-meta-MDD is a large-scale MDD research project with multi-data centers with the aim to advance the statistical capability and analytic heterogeneity of the data \cite{yan2019reduced}. Members of the consortium come from 25 research groups working in 17 hospitals in China \cite{yan2010dparsf} (DPARSF, http://rfmri.org). The consortium obtains rsfMRI indices of 1300 MDD patients and 1128 healthy controls and tabular clinical features including diagnosis, education, age, sex, episode status (if the patient’s prior and current episodes are diagnosed as MDD), medication status (whether antidepressants are used), and illness duration (months). The scanning parameters and demographic information for each site are provided in Supplementary Tables S1-S3. All patients are diagnosed according to the Diagnostic and Statistical Manual of Mental Disorders-IV. The severity of depression is rated using the Hamilton Depression Rating Scale. Deidentified and anonymized data are contributed from studies approved by local Institutional Review Boards. All study participants have provided written informed consent at their local institution.

\subsection{Data acquisition and preprocessing}
Data pre-processing steps include within-subject spatial realignment, head motion correction, slice timing correction, and spatial normalization \cite{yang2021disrupted}. Detrending and temporal bandpass filtering of the fMRI data are carried out to reduce the effects of low-frequency drifts and physiological high-frequency noise. For each subject, the first five volumes of the scanned data are removed for signal equilibrium. Realignment and re-slicing are applied to the remaining volumes to correct the head motion. The whole-brain functional network is obtained by applying Fisher’s r-to-z transform to the functional connectivity (FC) matrix of each subject \cite{gai2022classification}. We arrive at 1157 MDD patients and 1128 controls after excluding the subjects with null values. 

\subsection{Features considered} 
We consider the following four rsfMRI features that are commonly investigated and utilized in past studies. These feature values are readily available by standard data processing tools, allowing the method to be conveniently implemented.

{\bf ReHo}: Regional homogeneity (ReHo) is a voxel-wise measure for the similarity of the blood oxygen level-dependent signal within a particular brain region. Essentially, it assesses how similar the time series of a voxel is to its neighboring voxels. Fluctuations in ReHo are indicative of local abnormalities in brain activity that may be associated with mental disorders \cite{li2020identification}.

{\bf DC}: Degree centrality (DC) \cite{gao2016decreased,sacchet2015support} is a concept in network science, which quantifies the importance of a node in terms of its connectivity. It is suggested that mental illness may emerge when the connection among different brain regions becomes abnormal such that they do not cooperate harmoniously. Therefore, DC is believed to be a feature related to mental illness.

{\bf fALFF}: Amplitude of low-frequency fluctuations (ALFF) \cite{yu2007altered} measures the energy intensity of time series in the low-frequency band. The fMRI time series of each voxel is transformed into the frequency domain using a fast Fourier transform. The power spectrum is then computed for each voxel, and ALFF is obtained by summing the power within the predefined frequency range. The fractional amplitude of low-frequency fluctuations (fALFF) considered in this study is similar to ALFF. The difference is that fALFF calculates the ratio of the power within the low-frequency range (0.01-0.1 Hz) to the total power across the entire frequency spectrum. The normalization in fALFF makes it a better choice for comparisons between individuals. 

{\bf VMHC}: Voxel-mirrored homotopic connectivity (VMHC) \cite{stark2008regional,hermesdorf2016major} measures the similarity of the blood oxygen level-dependent signal between a pair of homotopic regions in the left and right hemispheres of the brain. The strength of these homotopic patterns can vary between regions \cite{savio2015local}. A higher VMHC value suggests stronger and more symmetric connectivity.

\subsection{The MFMC method}
\begin{figure}[htb!]
    \centering 
    \includegraphics[width=0.95\textwidth]{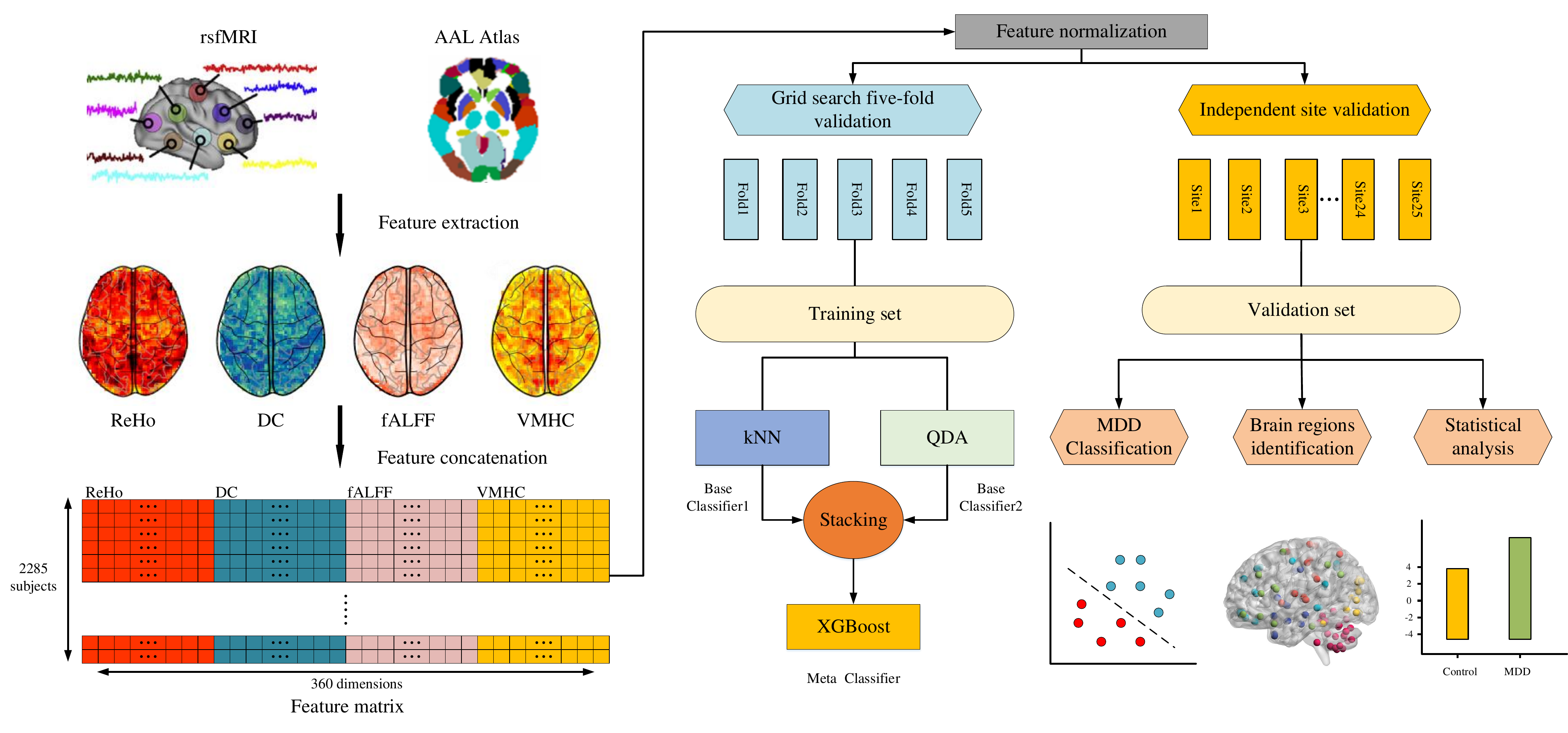} 
    \caption{
Schematic illustration of the MFMC for MDD discriminant analysis. The brain is divided into 90 brain regions. The value of ReHo, DC, fALFF, and VMHC are calculated for each brain region. These values are put together to form a 360 × 1 vector as the representation of a subject. The data is sent to the base classifier kNN and QDA. The result is further proceeded by the meta classier XGBoost.
}
    \label{overflow} 
\end{figure}

Fig. \ref{overflow} illustrates the flow of this study and the framework of MFMC. The brain parcellation is based on the automatic anatomical labeling (AAL) atlas, which gives rise to 90 brain regions of interest. The feature values of the brain regions are calculated using SPM12 and DPARSF toolbox under default parameter values. Each feature is represented by a  90 × 1 vector. The four feature vectors are combined to form a 360 × 1 vector for each individual. The concatenated feature vectors are used to train and validate the machine learning model. To reduce the impact of site effects, we use the combat method \cite{maikusa2021comparison} to normalize the data set before training. Subjects are randomly split into the training set and testing set, where the training set is used to build the classifier and the testing set is used to evaluate the performance. We use the stacking method \cite{liang2022multi,luo2022accelerated} to integrate three supervised learning classifiers that are k-Nearest Neighbor (kNN) \cite{zang2021effects}, quadratic discriminant analysis (QDA) \cite{movahed2021major}, and XGBoost \cite{shi2021multivariate}. kNN and QDA are the base classifiers whose outputs are used to train the meta classifier XGBoost. We conduct a grid search \cite{syarif2016svm} for each classifier to determine the optimal hyper-parameters. The specific process of the stacking method is shown in Supplementary Information S1. The algorithms are implemented through the Python-based sklearn package. The codes of MFMC are made publicly available at https://github.com/Luoyunsong/MFMC-MDD-discrimination.
 
\section{Results}
\subsection{Classification performance}

\begin{table}[hbt!]
\centering
\caption{The MDD discrimination accuracy by MFMC and other 13 related studies. We record the feature considered by the model and the classifier of the model. FC corresponds to functional connectivity and sMRI is structural magnetic resonance imaging. Both cross-validation and leave-one-out are considered by past studies. We record the reported accuracy in past studies.}
\label{studies}
\begin{threeparttable}
\resizebox{\linewidth}{!}{%
\begin{tblr}{
  vlines,
  cells = {c},
  cell{1}{1} = {r=2}{},
  cell{1}{2} = {c=2}{},
  cell{1}{4} = {r=2}{},
  cell{1}{5} = {r=2}{},
  cell{1}{6} = {c=2}{},
  cell{11}{1} = {r=2}{},
  cell{11}{2} = {r=2}{},
  cell{11}{3} = {r=2}{},
  cell{11}{5} = {r=2}{},
  cell{11}{6} = {r=2}{},
  cell{11}{7} = {r=2}{},
  cell{17}{1} = {r=2}{},
  cell{17}{2} = {r=2}{},
  cell{17}{3} = {r=2}{},
  cell{17}{5} = {r=2}{},
  cell{17}{6} = {r=2}{},
  cell{17}{7} = {r=2}{},
  hline{1,3-11,13-17,19} = {-}{},
  hline{2} = {2-3,6-7}{},
}
References             & Subjects &          & Feature                & Method   & Accuracy         &               \\
                       & MDDs     & Controls &                        &          & Cross-validation & Leave-one-out \\
Zhi D et al \cite{zhi2021bncpl}           & 208      & 210      & FC (rsfMRI)            & CNN      & 0.71             & 0.664         \\
Gao Y et al \cite{gao2017discriminating}           & 198      & 234      & fALFF (rsfMRI)         & SVM      & 0.738426         & —             \\
Wang Q et al $^1$ \cite{wang2022adaptive}     & 282      & 251      & rsfMRI \& sMRI             & GCN \& CNN   & 0.65             & —              \\
Yao D et al $^1$ \cite{yao2020temporal}      & 282      & 251      & FC (rsfMRI)            & GCN      & 0.738            & —         \\
Yao D et al $^1$ \cite{yao2021tensor}       & 282      & 251      & fALFF, VMHC \& DC       & TMRL     & 0.642            & —             \\
Fang Y et al $^1$ \cite{fang2023unsupervised}      & 356      & 325      & rsfMRI                 & GCN      & —                & 0.5973             \\
El-Gazzar A et al $^1$ \cite{el2022fmri} & 825      & 628      & FC (rsfMRI)            & CNN      & 0.654            & —             \\
Qin K  et al $^1$ \cite{qin2022using}       & 821      & 765      & FC (rsfMRI)            & GCN      & 0.815            & 0.813        \\
Shi Y et al $^1$ \cite{shi2021multivariate}        & 1021     & 1100     & FC (rsfMRI)            & XGBoost  & 0.728            & —             \\
                       &          &          & (rsfMRI)               &          &                  &               \\
Gao J et al $^1$ \cite{gao2023classification}      & 1300     & 1128     & sMRI                   & 3D CNN   & 0.7654           & —        \\
Liang Y et al $^1$ \cite{liang2022multi}    & 1300     & 1128     & FC (rsfMRI)            & DNN      & 0.654            & 0.634             \\
Nunes A et al $^2$ \cite{nunes2020using} & 853      & 2167     & sMRI                   & SVM      & —                & 0.6523         \\
Belov V et al $^2$ \cite{belov2022global} & 2288     & 3077     & sMRI                   & SVM      & 0.62             & —             \\
MFMC $^1$           & 1157     & 1128     & ReHo, fALFF, VMHC \& DC & Stacking & \bf{0.9688}           & \bf{0.9217}        \\
                       &          &          & (rsfMRI)               &          &                  &               
\end{tblr}
}
\begin{tablenotes}    
    \footnotesize               	  
    \item[1] Rest-meta-MDD data set
    \item[2] ENGIMA data set
\end{tablenotes} 
\end{threeparttable}
\end{table}

To comprehensively evaluate the classification performance of MFMC, we apply both 5-fold cross-validation and leave-one-out for model testing. The MDD discrimination accuracy is presented in Table 1 and other metrics such as specificity, sensitivity, and F1-score are listed in Supplementary Table S4. Overall, MFMC reaches a discrimination accuracy 96.88$\%$, with a sensitivity 96.40$\%$ and a specificity 97.38$\%$. To compare with other state-of-art methods, we collect 13 recent MDD discrimination studies whose sample sizes are close to ours. Unfortunately, many of these works do not have the codes publicly available, making it impossible to implement their methods on the same data with the same testing process. Therefore, we chose the reported performance of these methods as the baseline (Table \ref{studies}). Since the reported performance is usually the most optimal result for the data investigated, and our work considers more subjects than most of these studies, such a comparison is slightly against MFMC. But MFMC still outperforms others with a significant improvement. For example, compared with three studies on the Rest-meta-MDD data set with a similar number of subjects \cite{shi2021multivariate,liang2022multi,el2022fmri}, MFMC achieves over 20 percentage points advance in classification accuracy. This demonstrates the benefit of concatenating multiple features and stacking multiple classifiers in MDD discrimination.

To separately analyze the contribution of multi-feature concatenation and multi-classifier stacking, we report the classification performance by a single classifier with different combinations of features (Supplementary Tables S5-S7). When there is only one feature considered, the classification performance is always low, regardless of the classifier applied. As more features are concatenated, the classification accuracy grows drastically. In some cases, the use of only two features (e.g. ReHo and fALFF) can yield results close to the optimal. But the combination of the four features gives the best performance. When only one classifier is applied, QDA outperforms the other two classifiers. But the performance is improved further when stacking all three classifiers together. The use of XGBoost as the meta classifier is the best compared with the other two choices of mete classifier (Supplementary Tables S8-S9).

\subsection{Generalizability}
The generalizability of a machine learning algorithm refers to its ability to perform well on new data that it has not been previously exposed to. In clinical practices, an MDD discrimination model trained on existing data will face new subjects from other hospitals. Therefore, the generalizability of a model is an important factor towards its successful application. For this reason, we investigate the generalizability of MFMC by separating sites in the training and testing data. Different from the general 5-fold cross-validation where all subjects from different sites are mixed together, now subjects used to train and test the model are from distinct sites. This mimics the scenario in clinical application that the trained model is applied to diagnose new patients. The results are reported in Table \ref{independent} in which one site is used for testing and the rest data are used to train the model. The discrimination accuracy remains high, except for site 1. This phenomenon is in line with another study \cite{qin2022using}, which reports a much lower discrimination accuracy on site 1. We suspect this may be caused by different instruments used on that particular site that adopt different voxel size parameters. 

In addition, we note that the number of subjects on each individual site is usually smaller than that in the 5-fold cross-validation. As the discrimination accuracy tends to be higher on small data sets, the performance on one site may not fully reflect the generalizability. Therefore, we combine multiple sites to make the number of subjects in the testing set comparable with that in the 5-fold cross-validation. The results are reported in Supplementary Table S10. Compared with the performance under 5-fold cross-validation, the discrimination accuracy drops. But except in cases that contain site 1, the drop $\Delta_\text{accuracy}$ is not big. Overall, when applied to subjects from independent sites, MFMC identifies MDD patients more accurately than existing methods when applied to homogeneously mixed subjects. The performance advances support the good generalizability of MFMC.

\begin{table}[hbt!]
\centering
\caption{The performance of MFMC when subjects in the training and testing set are from independent sites.}
\label{independent}
\begin{tabular}{@{}ccccccc@{}}
\toprule\hline
Testing set & Accuracy (\%)& Sensitivity (\%)& Specificity (\%)& F1 (\%)   & NC  & MDD \\ \midrule\hline
Site1   & 66.22    & 33.78       & 98.65           & 50.00 & 74  & 74  \\
Site2   & 98.33    & 96.67       & 1           & 98.31 & 30  & 30  \\
Site3   & 94.59    & 0           & 94.59       & 0     & 37  & 0   \\
Site4   & 95.83    & 1       & 91.67       & 96 & 24  & 24  \\
Site5   & 1        & 0           & 1           & 0     & 11  & 0   \\
Site6   & 1        & 1           & 1           & 1     & 15  & 15  \\
Site7   & 98.85    & 1           & 97.96       & 98.70 & 49  & 38  \\
Site8   & 91.33    & 85.33       & 97.33       & 90.78 & 75  & 75  \\
Site9   & 86.00    & 0           & 86.00       & 0     & 50  & 0   \\
Site10  & 96.39    & 1           & 90.91       & 97.09 & 33  & 50  \\
Site11  & 1        & 1           & 1           & 1     & 29  & 8   \\
Site12  & 1        & 1           & 1           & 1     & 6   & 32  \\
Site13  & 1        & 1           & 1           & 1     & 17  & 25  \\
Site14  & 1        & 1           & 1           & 1     & 32  & 64  \\
Site15  & 1    & 1           & 1       & 1 & 50  & 50  \\
Site16  & 98.39        & 1           & 96.77           & 98.41     & 31  & 31  \\
Site17  & 97.80    & 95.74       & 1           & 97.83 & 44  & 47  \\
Site18  & 95.12    & 90.48       & 1           & 95.00 & 20  & 21  \\
Site19  & 1    & 1           & 1       & 1 & 36  & 51  \\
Site20  & 90.93    & 97.24       & 83.27       & 90.98 & 251 & 254 \\
Site21  & 89.03    & 97.65           & 78.57       & 90.71 & 70  & 85  \\
Site22  & 1        & 1           & 1           & 1     & 20  & 30  \\
Site23  & 1        & 1           & 1           & 1     & 30  & 32  \\
Site24  & 1    & 1           & 1       & 1 & 31  & 32  \\
Site25  & 98.03    & 98.88           & 96.83       & 98.32 & 63  & 89  \\ \bottomrule\hline
\end{tabular}
\end{table}

\subsection{The most discriminative features}

\begin{figure}[htb!]
    \centering
    \includegraphics[width=0.9\textwidth]{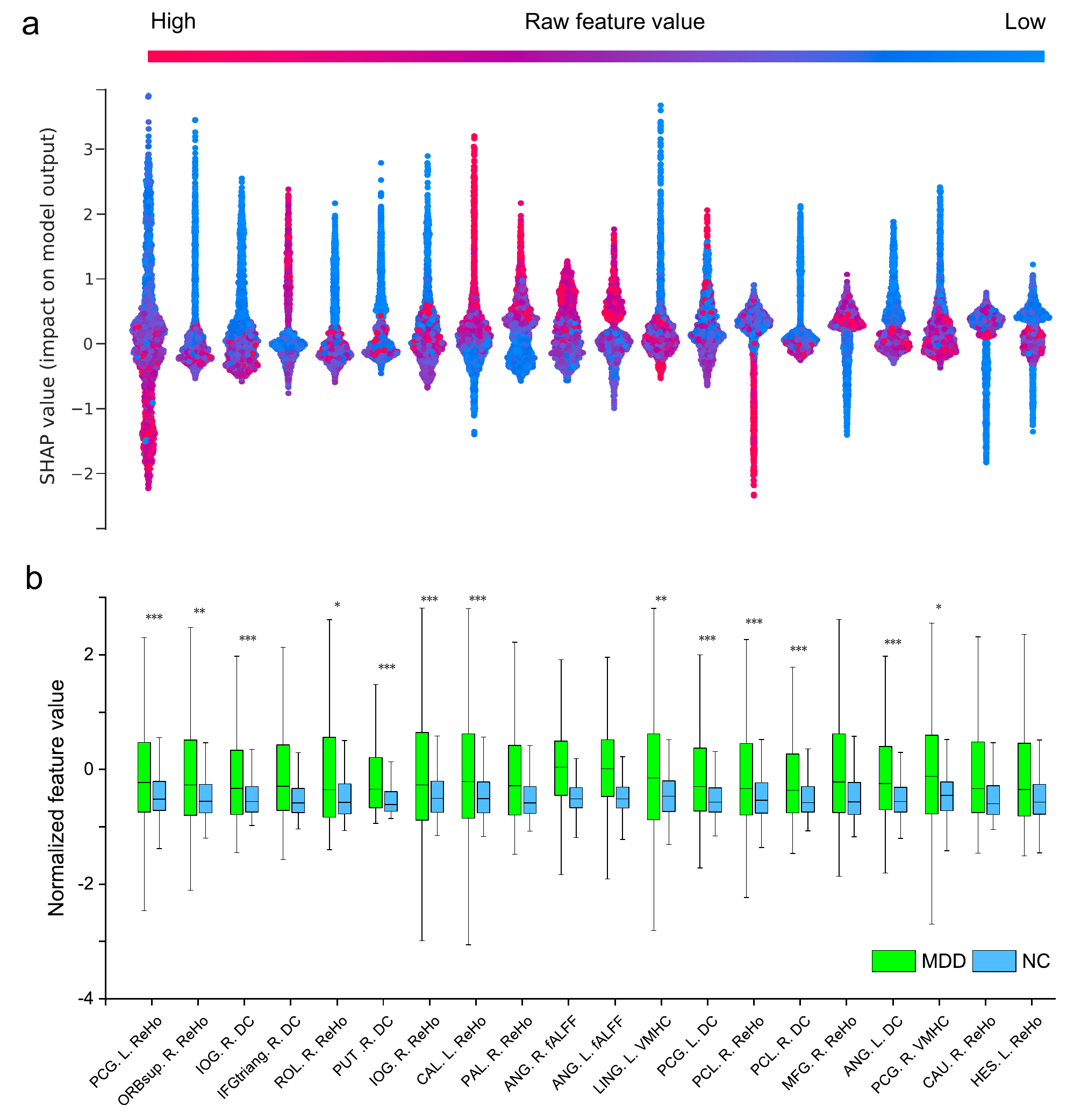} 
    \caption{The top 20 features ranked by their SHAP values. {\bf (a)}. Each dot corresponds to a subject. For a given feature, the horizontal axis records the subject's SHAP value and the color represents the subject's raw feature value. MFMC identifies a subject as MDD if the SHAP value is positive, and as normal control if the SHAP value is negative. {\bf (b)}. The normalized feature values of MDD patients and normal controls. The normalization is done by the z-score method. The two-sample t-test is performed to calculate the p-value. * represents $p < 0.05$, ** represents $p < 0.01$, and *** represents $p < 0.001$.}
    \label{SHAP} 
\end{figure}

\begin{figure}[htb!]
    \centering 
    \includegraphics[width=0.8\textwidth]{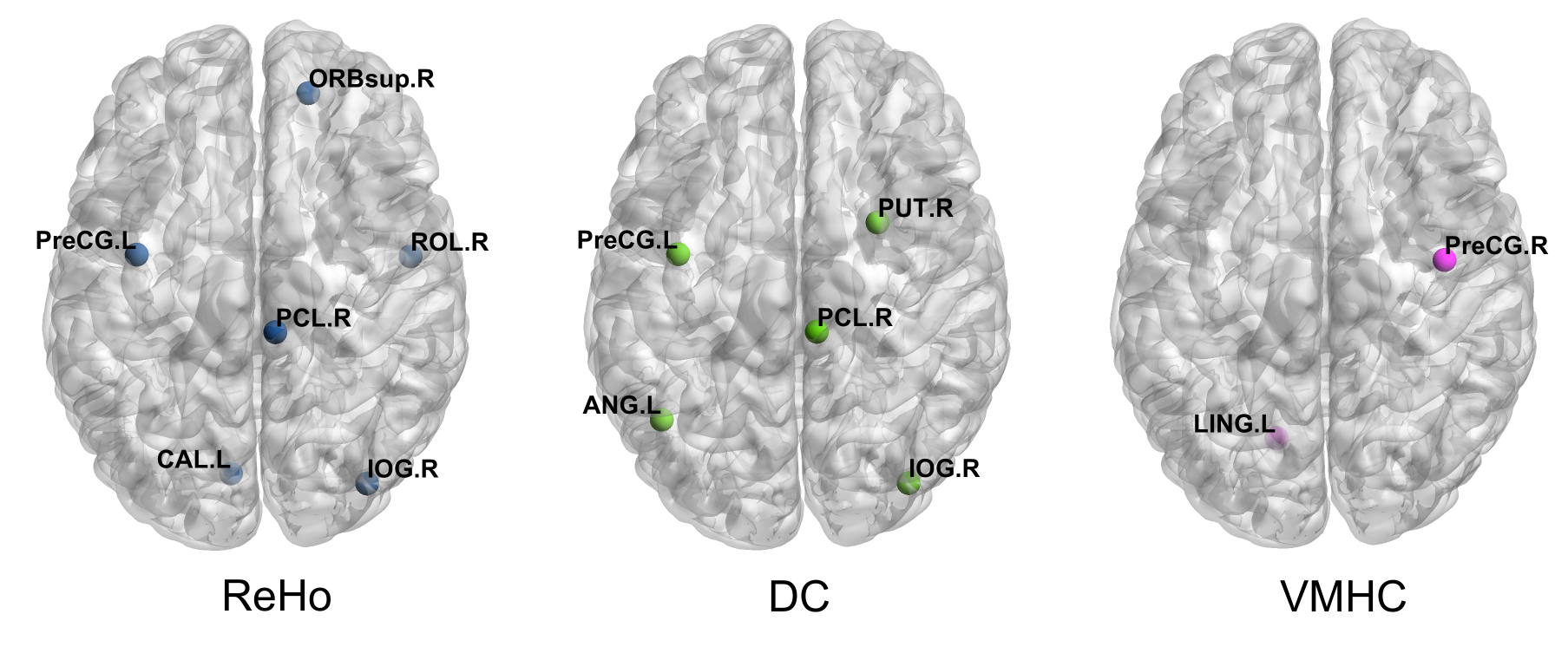} 
    \caption{The most discriminative brain regions. Among the top 20 features ranked by SHAP values, 13 features show significant differences at the group level, including posterior cingulate gyrus, orbital part of superior frontal gyrus, inferior occipital gyrus, rolandic operculum, lenticular nucleus putamen, calcarine fissure and surrounding cortex, lingual gyrus, angular gyrus, and paracentral lobule.}
    \label{Top 20} 
\end{figure}

The choice of XGBoost as the meta classifier allows us to probe the decision process of MFMC. With the SHAP value, we obtain features contributing most to MDD discrimination. Fig. \ref{SHAP}a shows the top 20 out of 360 features ranked by SHAP values. The use of these 20 features alone allows us to reach the MDD discrimination accuracy 87.0\%, which is roughly 90 percent of MFMC's full performance when taking all 360 feature values (Supplementary Table S14). For these 20 features, we further perform a two-sample t-test on the normalized feature values of MDD patients and normal controls. We narrow down to 13 features that are significantly different at the group level Fig. \ref{SHAP}b. The 13 features involve 9 brain regions, which are the posterior cingulate gyrus (PCG), orbital part of superior frontal gyrus (ORBsup), inferior occipital gyrus (IOG), rolandic operculum (ROL), lenticular nucleus putamen (PUT), calcarine fissure and surrounding cortex (CAL), lingual gyrus (LING), paracentral lobule (PCL), and angular gyrus (ANG). Among them, the posterior cingulate gyrus, inferior occipital gyrus, and paracentral lobule contain more than one feature, implying their importance in MDD identification. We also visualize the brain regions of the 13 features using the BrainNet viewer \cite{xia2013brainnet} (Fig. \ref{Top 20}). It is noteworthy that the 13 features can yield a discrimination accuracy 84.5\%, higher than the top 13 features ranked by their SHAP values (Supplementary Table S14). This suggests their potentials in distinguishing MDD patients while explaining the results with statistics at the group level.

\section{Discussion and conclusion}

In this study, we propose a method MFMC for MDD discrimination. The method is tested on the Rest-meta-MDD, a large MDD data set collected from multiple sites in China. By concatenating multiple rsfMRI features and stacking multiple classifiers, MFMC achieves promising discrimination accuracy with a significant improvement compared with other studies' reported performance. At the same time, MFMC demonstrates good generalizability. When applied to subjects from distinct sites, the accuracy of MFMC drops slightly but still outperforms other state-of-art methods. Finally, by combining SHAP values and statistical differences, we obtain 13 features associated with 9 brain regions. The use of these 13 features allows us to reach 87 percent of MFMC's full capacity when taking all 360 features into consideration.

On one hand, feature-level fusion integrates different types of features to convey more comprehensive and complementary information. Several studies have shown that the concatenation of multi-type features or modalities can considerably improve the classification performance in the diagnosis of psychiatric disorders \cite{huang2020fusion,dai2012discriminative,zhang2021alzheimer}. It can be found from Supplementary Tables S5-S7 that the accuracy of one single classifier dramatically improves through feature-level fusion. While the classification accuracy is higher for the same feature when decision-level fusion is applied (Supplementary Table S4), the improvement is relatively small. Therefore, both feature concatenation and model stacking contribute to the discrimination accuracy of MFMC, but feature-level fusion plays a substantially more important role. On the other hand, the decision-level fusion, which combines diverse classifiers to establish a more sophisticated decision function, can potentially enhance the generalizability of the model. This is verified by results in Supplementary Tables S10-S13, where the training and testing subjects are from independent sites and multiple sites are combined to make the number of subjects in the testing set similar to that under 5-fold cross-validation. The discrimination accuracy of the classifier drops compared with its performance under 5-fold cross-validation. But stacking multiple classifiers gives rise to less accuracy drop $\Delta_\text{accuracy}$ than applying one single classifier, which demonstrates the improved generalizability by decision-level fusion.

We choose XGBoost as the meta classifier not only because it gives the best performance compared with kNN and QDA (Supplementary Tables S8-S9), but also because XGBoost allows us to interpret the classification results with the SHAP value. By associating the SHAP value of a feature and its statistical difference between MDD patients and normal controls, we identify 13 features statistically different at the group level and also important in MDD discrimination. The results are consistent with some previous studies on ReHo \cite{yao2009regional,guo2011abnormal,kong2017electroconvulsive}, DC \cite{qin2014abnormal}, and VMHC \cite{wang2015interhemispheric,zhong2016functional}. In general, the ReHo feature accounts for most of the 13 features and the brain region posterior cingulate gyri \cite{yang2016decreased,yan2019reduced} is the main contributor in discriminating MDD from controls. Some other brain regions, such as the orbital part of superior frontal gyrus \cite{zhang2011disrupted}, inferior occipital gyrus \cite{li2022variability}, rolandic operculum \cite{sun2022comparative}, lenticular nucleus putamen \cite{su2014cerebral}, calcarine fissure and surrounding cortex \cite{zhang2011disrupted,yu2018functional}, lingual gyrus \cite{liu2017decreased,jung2014impact}, paracentral lobule \cite{zhang2021increased,jiang2020common}, and angular gyrus \cite{mo2020bifrontal}, have been investigated in previous studies and their roles in MDD are reported.

The task of identifying the group difference and predicting a single subject are quite different as they try to address distinct research questions \cite{calhoun2017prediction}. In general, showing group differences is much easier \cite{arbabshirani2017single}. The proposed method allows us to make predictions for individual subjects while providing some explanations. The feature values of brain regions we identified with significant differences are consistent with the findings of several previous studies at the group level. Compared to healthy controls, MDD patients show significantly decreased ReHo in the posterior cingulate gyrus \cite{yao2009regional,mo2020bifrontal} and increased ReHo in the calcarine fissure surrounding cortex \cite{shen2017identify}. In other words, our model is inclined to diagnose an individual as an MDD patient if the subject has a low ReHo in the posterior cingulate gyrus and a high Reho in the calcarine fissure surrounding cortex (Supplementary Information S2).

There are still some limitations in this study. Due to the limited data availability, we can only test MFMC on the REST-meta-MDD data set. The discrimination performance on other data remains untested. This gives rise to an intriguing question if MFMC can be effectively transferred. For this reason, we make the code publicly available for future studies. Moreover, only four features are adopted in this study. There are still many features from other modalities such as sMRI, and diffusion MRI, which all provide useful information about a subject \cite{cherubini2016importance,rokicki2021multimodal}. How to integrate multi-modal features to improve the model performance needs further investigation \cite{calhoun2016multimodal}.  Finally, depression is a long-term process. More finely delineation of the depression is required such that models can identify different stages of depression and provide guidance for prognostic treatment \cite{schmaal2017cortical}.

\section*{Funding and Disclosure}
This work is supported by the University Innovation Research Group of Chongqing (No. CXQT21005) and the Fundamental Research Funds for the Central Universities (No. SWU-XDJH202303). The authors declare no competing interests.

\section*{CRediT authorship contribution statement}
Yunsong Luo: Writing - Original Draft, Writing - Review \& Editing, Visualization, Methodology, Formal analysis. Wenyu Chen: Formal analysis. Ling Zhan: Formal analysis. Jiang Qiu: Resources. Tao Jia: Conceptualization, Supervision, Writing - Original Draft, Writing - Review \& Editing.

\section*{Declaration of Competing Interest}
The authors declare that they have no known competing financial interests or personal relationships that could have appeared to influence the work reported in this paper.

\section*{Acknowledgments}
We thank the other collaborative members of the REST-meta-MDD consortium for sharing the data.

\clearpage

\bibliographystyle{LYS}
\bibliography{mybibfile}

\begin{thebibliography}{10}

\bibitem{murray2012disability}
Murray C~J, Vos T, Lozano R, Naghavi M, Flaxman A~D, Michaud C, et~al.
\newblock Disability-adjusted life years (dalys) for 291 diseases and injuries
  in 21 regions, 1990--2010: a systematic analysis for the global burden of
  disease study 2010.
\newblock The lancet, 2012; 380:2197--2223.

\bibitem{gupta2019machine}
Gupta S, Vig R.
\newblock Machine learning models for depression patient classification using
  fmri: A study.
\newblock In International Conference on Computational Science and Its
  Applications. Springer, 2019;  685--696.

\bibitem{schnack2016detecting}
Schnack H~G, Kahn R~S.
\newblock Detecting neuroimaging biomarkers for psychiatric disorders: sample
  size matters.
\newblock Frontiers in psychiatry, 2016; 7:50.

\bibitem{qin2022using}
Qin K, Lei D, Pinaya W~H, Pan N, Li W, Zhu Z, et~al.
\newblock Using graph convolutional network to characterize individuals with
  major depressive disorder across multiple imaging sites.
\newblock EBioMedicine, 2022; 78:103977.

\bibitem{zatorre2012plasticity}
Zatorre R~J, Fields R~D, Johansen-Berg H.
\newblock Plasticity in gray and white: neuroimaging changes in brain structure
  during learning.
\newblock Nature neuroscience, 2012; 15:528--536.

\bibitem{gao2018machine}
Gao S, Calhoun V~D, Sui J.
\newblock Machine learning in major depression: From classification to
  treatment outcome prediction.
\newblock CNS neuroscience \& therapeutics, 2018; 24:1037--1052.

\bibitem{nouretdinov2011machine}
Nouretdinov I, Costafreda S~G, Gammerman A, Chervonenkis A, Vovk V, Vapnik V,
  et~al.
\newblock Machine learning classification with confidence: application of
  transductive conformal predictors to mri-based diagnostic and prognostic
  markers in depression.
\newblock Neuroimage, 2011; 56:809--813.

\bibitem{zhang2016discriminative}
Zhang Q, Wu Q, Zhang J, He L, Huang J, Zhang J, et~al.
\newblock Discriminative analysis of migraine without aura: using functional
  and structural mri with a multi-feature classification approach.
\newblock PloS one, 2016; 11:e0163875.

\bibitem{guo2019resting}
Guo H, Li Y, Mensah G~K, Xu Y, Chen J, Xiang J, et~al.
\newblock Resting-state functional network scale effects and statistical
  significance-based feature selection in machine learning classification.
\newblock Computational and Mathematical Methods in Medicine, 2019; 2019.

\bibitem{yan2020quantitative}
Yan B, Xu X, Liu M, Zheng K, Liu J, Li J, et~al.
\newblock Quantitative identification of major depression based on
  resting-state dynamic functional connectivity: a machine learning approach.
\newblock Frontiers in neuroscience, 2020; 14:191.

\bibitem{jin2020region}
Jin J, Huang L.
\newblock A region-based feature extraction method for rs-fmri of depressive
  disorder classification.
\newblock In 2020 International Conference on Computer Vision, Image and Deep
  Learning (CVIDL). IEEE, 2020;  707--710.

\bibitem{shimizu2020maximum}
Shimizu Y, Yoshimoto J, Takamura M, Okada G, Matsumoto T, Fuchikami M, et~al.
\newblock Maximum credibility voting (mcv) an integrative approach for accurate
  diagnosis of major depressive disorder from clinically readily available
  data.
\newblock In 2020 Asia-Pacific Signal and Information Processing Association
  Annual Summit and Conference (APSIPA ASC). IEEE, 2020;  1023--1032.

\bibitem{wei2013identifying}
Wei M, Qin J, Yan R, Li H, Yao Z, Lu Q.
\newblock Identifying major depressive disorder using hurst exponent of
  resting-state brain networks.
\newblock Psychiatry Research: Neuroimaging, 2013; 214:306--312.

\bibitem{lundberg2020local}
Lundberg S~M, Erion G, Chen H, DeGrave A, Prutkin J~M, Nair B, et~al.
\newblock From local explanations to global understanding with explainable ai
  for trees.
\newblock Nature machine intelligence, 2020; 2:56--67.

\bibitem{zhang2020clinically}
Zhang K, Liu X, Shen J, Li Z, Sang Y, Wu X, et~al.
\newblock Clinically applicable ai system for accurate diagnosis, quantitative
  measurements, and prognosis of covid-19 pneumonia using computed tomography.
\newblock Cell, 2020; 181:1423--1433.

\bibitem{yan2019reduced}
Yan C~G, Chen X, Li L, Castellanos F~X, Bai T~J, Bo Q~J, et~al.
\newblock Reduced default mode network functional connectivity in patients with
  recurrent major depressive disorder.
\newblock Proceedings of the National Academy of Sciences, 2019;
  116:9078--9083.

\bibitem{yan2010dparsf}
Yan C, Zang Y.
\newblock Dparsf: a matlab toolbox for" pipeline" data analysis of
  resting-state fmri.
\newblock Frontiers in systems neuroscience, 2010; 4:13.

\bibitem{yang2021disrupted}
Yang H, Chen X, Chen Z~B, Li L, Li X~Y, Castellanos F~X, et~al.
\newblock Disrupted intrinsic functional brain topology in patients with major
  depressive disorder.
\newblock Molecular psychiatry, 2021; 26:7363--7371.

\bibitem{gai2022classification}
Gai Q, Chu T, Che K, Li Y, Dong F, Zhang H, et~al.
\newblock Classification of major depressive disorder based on integrated
  temporal and spatial functional mri variability features of dynamic brain
  network.
\newblock Journal of Magnetic Resonance Imaging, 2022; .

\bibitem{li2020identification}
Li H, Cui L, Cao L, Zhang Y, Liu Y, Deng W, et~al.
\newblock Identification of bipolar disorder using a combination of
  multimodality magnetic resonance imaging and machine learning techniques.
\newblock BMC psychiatry, 2020; 20:1--12.

\bibitem{gao2016decreased}
Gao C, Wenhua L, Liu Y, Ruan X, Chen X, Liu L, et~al.
\newblock Decreased subcortical and increased cortical degree centrality in a
  nonclinical college student sample with subclinical depressive symptoms: a
  resting-state fmri study.
\newblock Frontiers in human neuroscience, 2016; 10:617.

\bibitem{sacchet2015support}
Sacchet M~D, Prasad G, Foland-Ross L~C, Thompson P~M, Gotlib I~H.
\newblock Support vector machine classification of major depressive disorder
  using diffusion-weighted neuroimaging and graph theory.
\newblock Frontiers in psychiatry, 2015; 6:21.

\bibitem{yu2007altered}
Yu-Feng Z, Yong H, Chao-Zhe Z, Qing-Jiu C, Man-Qiu S, Meng L, et~al.
\newblock Altered baseline brain activity in children with adhd revealed by
  resting-state functional mri.
\newblock Brain and Development, 2007; 29:83--91.

\bibitem{stark2008regional}
Stark D~E, Margulies D~S, Shehzad Z~E, Reiss P, Kelly A~C, Uddin L~Q, et~al.
\newblock Regional variation in interhemispheric coordination of intrinsic
  hemodynamic fluctuations.
\newblock Journal of Neuroscience, 2008; 28:13754--13764.

\bibitem{hermesdorf2016major}
Hermesdorf M, Sundermann B, Feder S, Schwindt W, Minnerup J, Arolt V, et~al.
\newblock Major depressive disorder: findings of reduced homotopic connectivity
  and investigation of underlying structural mechanisms.
\newblock Human brain mapping, 2016; 37:1209--1217.

\bibitem{savio2015local}
Savio A, Gra{\~n}a M.
\newblock Local activity features for computer aided diagnosis of schizophrenia
  on resting-state fmri.
\newblock Neurocomputing, 2015; 164:154--161.

\bibitem{maikusa2021comparison}
Maikusa N, Zhu Y, Uematsu A, Yamashita A, Saotome K, Okada N, et~al.
\newblock Comparison of traveling-subject and combat harmonization methods for
  assessing structural brain characteristics.
\newblock Human brain mapping, 2021; 42:5278--5287.

\bibitem{liang2022multi}
Liang Y, Xu G.
\newblock Multi-level functional connectivity fusion classification framework
  for brain disease diagnosis.
\newblock IEEE Journal of Biomedical and Health Informatics, 2022; .

\bibitem{luo2022accelerated}
Luo Y, Chen W, Qiu J, Jia T.
\newblock Accelerated functional brain aging in major depressive disorder:
  evidence from a large scale fmri analysis of chinese participants.
\newblock Translational Psychiatry, 2022; 12:397.

\bibitem{zang2021effects}
Zang J, Huang Y, Kong L, Lei B, Ke P, Li H, et~al.
\newblock Effects of brain atlases and machine learning methods on the
  discrimination of schizophrenia patients: a multimodal mri study.
\newblock Frontiers in neuroscience, 2021; 944.

\bibitem{movahed2021major}
Movahed R~A, Jahromi G~P, Shahyad S, Meftahi G~H.
\newblock A major depressive disorder classification framework based on eeg
  signals using statistical, spectral, wavelet, functional connectivity, and
  nonlinear analysis.
\newblock Journal of Neuroscience Methods, 2021; 358:109209.

\bibitem{shi2021multivariate}
Shi Y, Zhang L, Wang Z, Lu X, Wang T, Zhou D, et~al.
\newblock Multivariate machine learning analyses in identification of major
  depressive disorder using resting-state functional connectivity: A
  multicentral study.
\newblock ACS Chemical Neuroscience, 2021; 12:2878--2886.

\bibitem{syarif2016svm}
Syarif I, Prugel-Bennett A, Wills G.
\newblock Svm parameter optimization using grid search and genetic algorithm to
  improve classification performance.
\newblock TELKOMNIKA (Telecommunication Computing Electronics and Control),
  2016; 14:1502--1509.

\bibitem{zhi2021bncpl}
Zhi D, Calhoun V~D, Wang C, Li X, Ma X, Lv L, et~al.
\newblock Bncpl: Brain-network-based convolutional prototype learning for
  discriminating depressive disorders.
\newblock In 2021 43rd Annual International Conference of the IEEE Engineering
  in Medicine \& Biology Society (EMBC). IEEE, 2021;  1622--1626.

\bibitem{gao2017discriminating}
Gao S, Osuch E~A, Wammes M, Th{\'e}berge J, Jiang T~Z, Calhoun V~D, et~al.
\newblock Discriminating bipolar disorder from major depression based on kernel
  svm using functional independent components.
\newblock In 2017 IEEE 27th international workshop on machine learning for
  signal processing (MLSP). IEEE, 2017;  1--6.

\bibitem{wang2022adaptive}
Wang Q, Li L, Qiao L, Liu M.
\newblock Adaptive multimodal neuroimage integration for major depression
  disorder detection.
\newblock Frontiers in Neuroinformatics, 2022; 16.

\bibitem{yao2020temporal}
Yao D, Sui J, Yang E, Yap P~T, Shen D, Liu M.
\newblock Temporal-adaptive graph convolutional network for automated
  identification of major depressive disorder using resting-state fmri.
\newblock In Machine Learning in Medical Imaging: 11th International Workshop,
  MLMI 2020, Held in Conjunction with MICCAI 2020, Lima, Peru, October 4, 2020,
  Proceedings 11. Springer, 2020;  1--10.

\bibitem{yao2021tensor}
Yao D, Yang E, Guan H, Sui J, Zhang Z, Liu M.
\newblock Tensor-based multi-index representation learning for major depression
  disorder detection with resting-state fmri.
\newblock In Medical Image Computing and Computer Assisted Intervention--MICCAI
  2021: 24th International Conference, Strasbourg, France, September
  27--October 1, 2021, Proceedings, Part V 24. Springer, 2021;  174--184.

\bibitem{fang2023unsupervised}
Fang Y, Wang M, Potter G~G, Liu M.
\newblock Unsupervised cross-domain functional mri adaptation for automated
  major depressive disorder identification.
\newblock Medical Image Analysis, 2023; 84:102707.

\bibitem{el2022fmri}
El-Gazzar A, Thomas R~M, Van~Wingen G.
\newblock fmri-s4: learning short-and long-range dynamic fmri dependencies
  using 1d convolutions and state space models.
\newblock In Machine Learning in Clinical Neuroimaging: 5th International
  Workshop, MLCN 2022, Held in Conjunction with MICCAI 2022, Singapore,
  September 18, 2022, Proceedings. Springer, 2022;  158--168.

\bibitem{gao2023classification}
Gao J, Chen M, Xiao D, Li Y, Zhu S, Li Y, et~al.
\newblock Classification of major depressive disorder using an attention-guided
  unified deep convolutional neural network and individual structural
  covariance network.
\newblock Cerebral Cortex, 2023; 33:2415--2425.

\bibitem{nunes2020using}
Nunes A, Schnack H~G, Ching C~R, Agartz I, Akudjedu T~N, Alda M, et~al.
\newblock Using structural mri to identify bipolar disorders--13 site machine
  learning study in 3020 individuals from the enigma bipolar disorders working
  group.
\newblock Molecular psychiatry, 2020; 25:2130--2143.

\bibitem{belov2022global}
Belov V, Erwin-Grabner T, Gonul A~S, Amod A~R, Ojha A, Aleman A, et~al.
\newblock Global multi-site benchmark classification of major depressive
  disorder using machine learning on cortical and subcortical features of 5,365
  participants from the enigma mdd dataset.
\newblock arXiv preprint arXiv:220608122, 2022; .

\bibitem{xia2013brainnet}
Xia M, Wang J, He Y.
\newblock Brainnet viewer: a network visualization tool for human brain
  connectomics.
\newblock PloS one, 2013; 8:e68910.

\bibitem{huang2020fusion}
Huang S~C, Pareek A, Seyyedi S, Banerjee I, Lungren M~P.
\newblock Fusion of medical imaging and electronic health records using deep
  learning: a systematic review and implementation guidelines.
\newblock NPJ digital medicine, 2020; 3:1--9.

\bibitem{dai2012discriminative}
Dai Z, Yan C, Wang Z, Wang J, Xia M, Li K, et~al.
\newblock Discriminative analysis of early alzheimer's disease using
  multi-modal imaging and multi-level characterization with multi-classifier
  (m3).
\newblock Neuroimage, 2012; 59:2187--2195.

\bibitem{zhang2021alzheimer}
Zhang Y, Wang S, Xia K, Jiang Y, Qian P, Initiative A~D~N, et~al.
\newblock Alzheimer’s disease multiclass diagnosis via multimodal
  neuroimaging embedding feature selection and fusion.
\newblock Information Fusion, 2021; 66:170--183.

\bibitem{yao2009regional}
Yao Z, Wang L, Lu Q, Liu H, Teng G.
\newblock Regional homogeneity in depression and its relationship with separate
  depressive symptom clusters: a resting-state fmri study.
\newblock Journal of affective disorders, 2009; 115:430--438.

\bibitem{guo2011abnormal}
Guo W~b, Liu F, Xue Z~m, Yu Y, Ma C~q, Tan C~l, et~al.
\newblock Abnormal neural activities in first-episode, treatment-naive,
  short-illness-duration, and treatment-response patients with major depressive
  disorder: a resting-state fmri study.
\newblock Journal of Affective Disorders, 2011; 135:326--331.

\bibitem{kong2017electroconvulsive}
Kong X~m, Xu S~x, Sun Y, Wang K~y, Wang C, Zhang J, et~al.
\newblock Electroconvulsive therapy changes the regional resting state function
  measured by regional homogeneity (reho) and amplitude of low frequency
  fluctuations (alff) in elderly major depressive disorder patients: an
  exploratory study.
\newblock Psychiatry Research: Neuroimaging, 2017; 264:13--21.

\bibitem{qin2014abnormal}
Qin J, Wei M, Liu H, Yan R, Luo G, Yao Z, et~al.
\newblock Abnormal brain anatomical topological organization of the
  cognitive-emotional and the frontoparietal circuitry in major depressive
  disorder.
\newblock Magnetic resonance in medicine, 2014; 72:1397--1407.

\bibitem{wang2015interhemispheric}
Wang Y, Zhong S, Jia Y, Zhou Z, Wang B, Pan J, et~al.
\newblock Interhemispheric resting state functional connectivity abnormalities
  in unipolar depression and bipolar depression.
\newblock Bipolar disorders, 2015; 17:486--495.

\bibitem{zhong2016functional}
Zhong X, Pu W, Yao S.
\newblock Functional alterations of fronto-limbic circuit and default mode
  network systems in first-episode, drug-na{\"\i}ve patients with major
  depressive disorder: a meta-analysis of resting-state fmri data.
\newblock Journal of affective disorders, 2016; 206:280--286.

\bibitem{yang2016decreased}
Yang R, Gao C, Wu X, Yang J, Li S, Cheng H.
\newblock Decreased functional connectivity to posterior cingulate cortex in
  major depressive disorder.
\newblock Psychiatry Research: Neuroimaging, 2016; 255:15--23.

\bibitem{zhang2011disrupted}
Zhang J, Wang J, Wu Q, Kuang W, Huang X, He Y, et~al.
\newblock Disrupted brain connectivity networks in drug-naive, first-episode
  major depressive disorder.
\newblock Biological psychiatry, 2011; 70:334--342.

\bibitem{li2022variability}
Li W, Wang C, Lan X, Fu L, Zhang F, Ye Y, et~al.
\newblock Variability and concordance among indices of brain activity in major
  depressive disorder with suicidal ideation: A temporal dynamics resting-state
  fmri analysis.
\newblock Journal of Affective Disorders, 2022; 319:70--78.

\bibitem{sun2022comparative}
Sun J~f, Chen L~m, He J~k, Wang Z, Guo C~l, Ma Y, et~al.
\newblock A comparative study of regional homogeneity of resting-state fmri
  between the early-onset and late-onset recurrent depression in adults.
\newblock Frontiers in psychology, 2022; 13:849847.

\bibitem{su2014cerebral}
Su L, Cai Y, Xu Y, Dutt A, Shi S, Bramon E.
\newblock Cerebral metabolism in major depressive disorder: a voxel-based
  meta-analysis of positron emission tomography studies.
\newblock BMC psychiatry, 2014; 14:1--7.

\bibitem{yu2018functional}
Yu H, Li F, Wu T, Li R, Yao L, Wang C, et~al.
\newblock Functional brain abnormalities in major depressive disorder using the
  hilbert-huang transform.
\newblock Brain imaging and behavior, 2018; 12:1556--1568.

\bibitem{liu2017decreased}
Liu X, Chen W, Hou H, Chen X, Zhang J, Liu J, et~al.
\newblock Decreased functional connectivity between the dorsal anterior
  cingulate cortex and lingual gyrus in alzheimer's disease patients with
  depression.
\newblock Behavioural brain research, 2017; 326:132--138.

\bibitem{jung2014impact}
Jung J, Kang J, Won E, Nam K, Lee M~S, Tae W~S, et~al.
\newblock Impact of lingual gyrus volume on antidepressant response and
  neurocognitive functions in major depressive disorder: a voxel-based
  morphometry study.
\newblock Journal of affective disorders, 2014; 169:179--187.

\bibitem{zhang2021increased}
Zhang R, Zhang L, Wei S, Wang P, Jiang X, Tang Y, et~al.
\newblock Increased amygdala-paracentral lobule/precuneus functional
  connectivity associated with patients with mood disorder and suicidal
  behavior.
\newblock Frontiers in human neuroscience, 2021; 14:585664.

\bibitem{jiang2020common}
Jiang X, Fu S, Yin Z, Kang J, Wang X, Zhou Y, et~al.
\newblock Common and distinct neural activities in frontoparietal network in
  first-episode bipolar disorder and major depressive disorder: Preliminary
  findings from a follow-up resting state fmri study.
\newblock Journal of affective disorders, 2020; 260:653--659.

\bibitem{mo2020bifrontal}
Mo Y, Wei Q, Bai T, Zhang T, Lv H, Zhang L, et~al.
\newblock Bifrontal electroconvulsive therapy changed regional homogeneity and
  functional connectivity of left angular gyrus in major depressive disorder.
\newblock Psychiatry Research, 2020; 294:113461.

\bibitem{calhoun2017prediction}
Calhoun V~D, Lawrie S~M, Mourao-Miranda J, Stephan K~E.
\newblock Prediction of individual differences from neuroimaging data.
\newblock Neuroimage, 2017; 145:135.

\bibitem{arbabshirani2017single}
Arbabshirani M~R, Plis S, Sui J, Calhoun V~D.
\newblock Single subject prediction of brain disorders in neuroimaging:
  Promises and pitfalls.
\newblock Neuroimage, 2017; 145:137--165.

\bibitem{shen2017identify}
Shen Z, Jiang L, Yang S, Ye J, Dai N, Liu X, et~al.
\newblock Identify changes of brain regional homogeneity in early and later
  adult onset patients with first-episode depression using resting-state fmri.
\newblock PloS one, 2017; 12:e0184712.

\bibitem{cherubini2016importance}
Cherubini A, Caligiuri M~E, P{\'e}ran P, Sabatini U, Cosentino C, Amato F.
\newblock Importance of multimodal mri in characterizing brain tissue and its
  potential application for individual age prediction.
\newblock IEEE journal of biomedical and health informatics, 2016;
  20:1232--1239.

\bibitem{rokicki2021multimodal}
Rokicki J, Wolfers T, Nordh{\o}y W, Tesli N, Quintana D~S, Aln{\ae}s D, et~al.
\newblock Multimodal imaging improves brain age prediction and reveals distinct
  abnormalities in patients with psychiatric and neurological disorders.
\newblock Human brain mapping, 2021; 42:1714--1726.

\bibitem{calhoun2016multimodal}
Calhoun V~D, Sui J.
\newblock Multimodal fusion of brain imaging data: a key to finding the missing
  link (s) in complex mental illness.
\newblock Biological psychiatry: cognitive neuroscience and neuroimaging, 2016;
  1:230--244.

\bibitem{schmaal2017cortical}
Schmaal L, Hibar D, S{\"a}mann P~G, Hall G, Baune B, Jahanshad N, et~al.
\newblock Cortical abnormalities in adults and adolescents with major
  depression based on brain scans from 20 cohorts worldwide in the enigma major
  depressive disorder working group.
\newblock Molecular psychiatry, 2017; 22:900--909.

\bibitem{shamaei2016suspended}
Shamaei E, Kaedi M.
\newblock Suspended sediment concentration estimation by stacking the genetic
  programming and neuro-fuzzy predictions.
\newblock Applied Soft Computing, 2016; 45:187--196.

\bibitem{lundberg2017unified}
Lundberg S~M, Lee S~I.
\newblock A unified approach to interpreting model predictions.
\newblock In Proceedings of the 31st international conference on neural
  information processing systems. 2017;  4768--4777.

\end{thebibliography}
\clearpage

\section*{Supplementary Information}

\subsection*{S1. Stacking model}

\begin{figure}[htb!]
\renewcommand{\thefigure}{S1}
\centering
\includegraphics[width=1\textwidth]{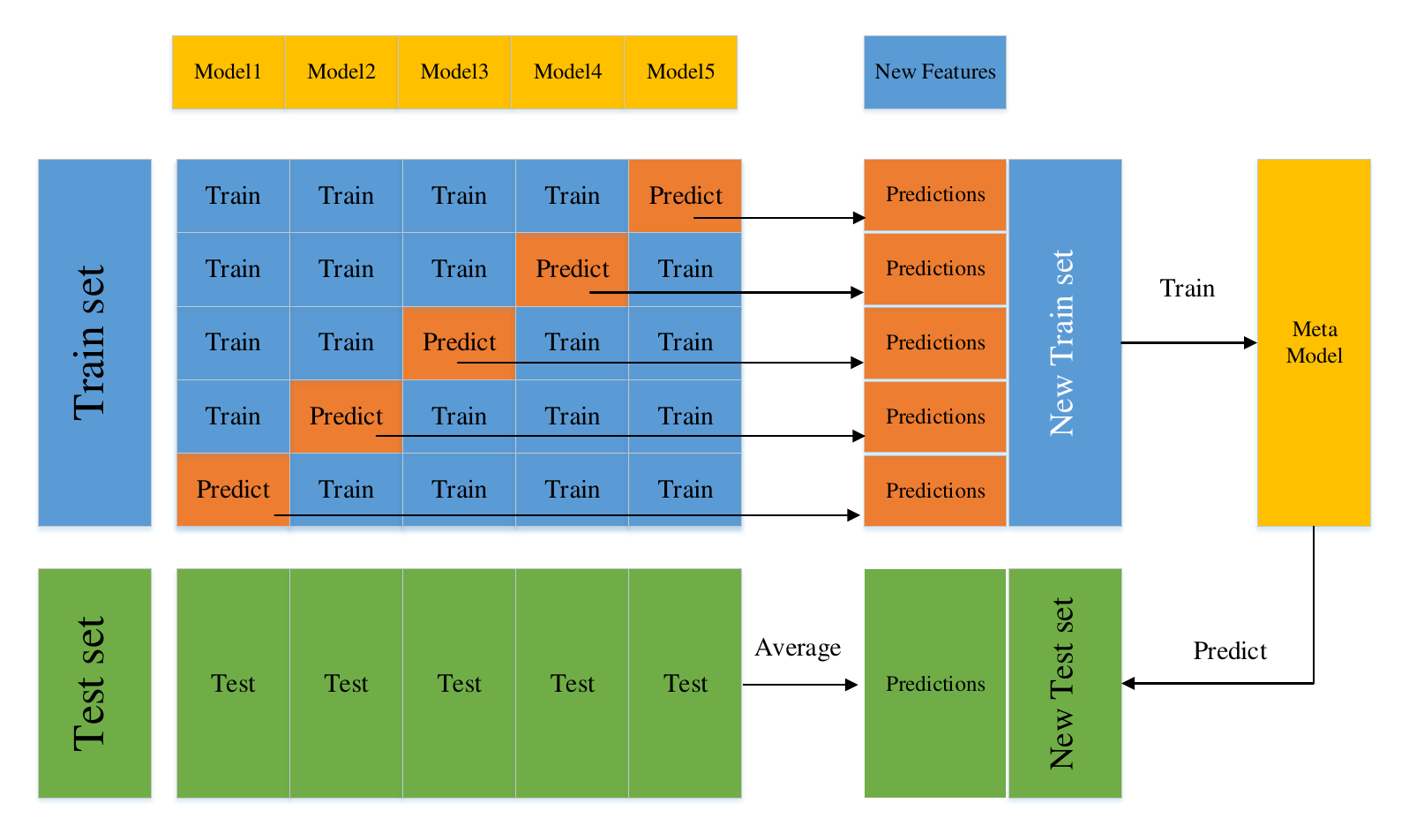}
\caption{Schematic diagram of two levels stacking.}
\label{stacking}
\end{figure}

Stacking model is an ensemble learning approach that combines different prediction models in a single model, working at levels or layers. This approach aims to minimize the errors of generalization by reducing the bias of its generalizers. Considering a stacking approach using two levels (level-0 and level-1) in Fig. \ref{stacking}. In level-0, diverse base models are trained, and the prediction of the response variable for each one is performed subsequently. These forecasts are used as a new input feature for the level-1 model, which is also called meta-model \cite{shamaei2016suspended}. At the same time, the base classifiers need to predict the test set, and the predictions are averaged as a new test set for the meta classifier to predict the desired result.

\clearpage

\subsection*{S2. SHAP Value analysis}
To understand which brain regions contribute to discriminating MDD patients from the perspective of disease diagnosis, we introduce the SHAP value \cite{lundberg2017unified,lundberg2020local,zhang2020clinically} to explain our model. SHAP values are represented as an additive feature attribution  method, it interprets the predicted value of the model as the sum of the imputed values of each feature in a linear model:

\begin{equation}
g\left(x^{\prime}\right)=\phi_{0}+\sum_{j=1}^{M} \phi_{j}
\label{Equation 1}
\end{equation}

where $g$ is the model to be interpreted, $M$ is the number of input features, the $\phi_{0}$ is a constant, and $\phi_{j} \in \mathbb{R}$ is the feature attribution for a feature $j$. The specific definition of $\phi_{j}$ is as follows:

\begin{equation}
\phi_{i}=\sum_{S \subseteq N \backslash\{i\}} \frac{|S| !(M-|S|-1) !}{M !}\left[f_{x}(S \cup\{i\})-f_{x}(S)\right]
\label{Equation 2}
\end{equation}

where $N$ is the set of all input features, $S$ is the set of non-zero indexes in $z^{\prime}$, $f_{x}(S)$ is the expected value of the function conditioned on a subset $S$ of the input features,  $\frac{|S| !(M-|S|-1) !}{M !}$ is the weight of the difference between the sample values under the condition of corresponding feature subset $S$, including feature $i$.

\begin{figure}[htb!]
    \renewcommand{\thefigure}{S2}
    \centering 
    \includegraphics[width=1\textwidth]{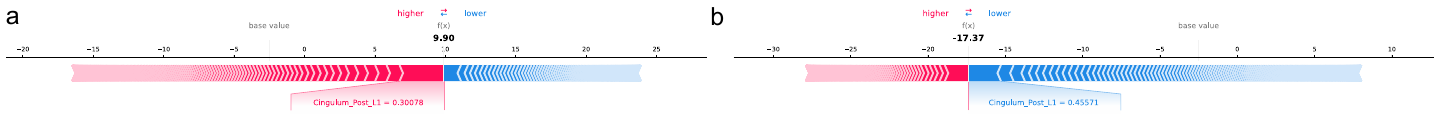} 
    \caption{SHAP dependency analysis}
    \label{SHAP dependency analysis} 
\end{figure}

\begin{figure}[htb!]
    \renewcommand{\thefigure}{S3}
    \centering 
    \includegraphics[width=1\textwidth]{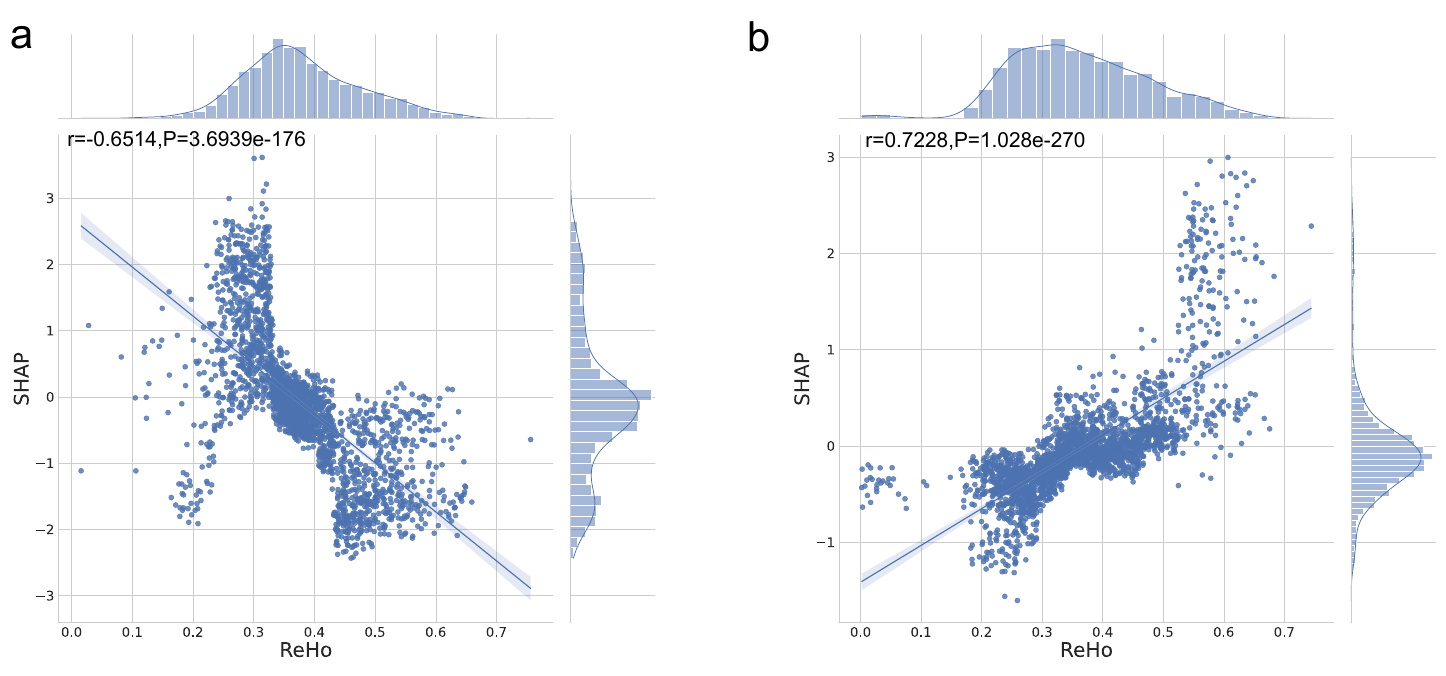} 
    \caption{Correlation between the ReHo value and SHAP value}
    \label{Correlation between the ReHo value and SHAP value} 
\end{figure}

MFMC is able to make predictions for individual subjects while also providing some interpretation. We select two subjects from the MDD and healthy control groups to show the effects of different features as the input risk factors for prognosis prediction. In Fig. \ref{SHAP dependency analysis}, there is a base SHAP value -2.493 to distinguish whether a subject is depressed or not. Pink features pushed the risk higher (to the right) and blue features pushed the risk lower (to the left). According to formula \ref{Equation 2}, if the cumulative SHAP value of each feature from a subject is greater than -2.493, then the subject will be identified as an MDD patient by our method. Otherwise, it will be identified as a normal control. Among the features of the two subjects, we can see that the value of the CingulumpostL1 provides the greatest contribution to the algorithm's judgment, that is, the CingulumpostL1=0.3008 is the longest red segment in Fig. \ref{SHAP dependency analysis}a, while the CingulumpostL1=0.4557 is the longest blue segment in Fig. \ref{SHAP dependency analysis}b. This is consistent with our previous conclusion that the Cingulumpost is a crucial feature to distinguish MDD from NC, and the smaller its value, the more likely it is to be identified as a MDD patient.

\begin{sidewaystable}[hbt!]
\renewcommand{\thetable}{S1}
\centering
\caption{Samples of selected sites.}
\label{Samples of selected sites}
\begin{tabular}{@{}llll@{}}
\hline
\toprule
Serial number & Research groups                                                                                                                                                           & MDD patients (n) & NCs (n) \\ \midrule \hline
1              & \begin{tabular}[c]{@{}l@{}}National Clinical Research Center for Mental Disorders, Peking University\end{tabular} & 74               & 74      \\
2              & \begin{tabular}[c]{@{}l@{}}The Affiliated Guangji Hospital of Soochow University\end{tabular}            & 30               & 30      \\
3              & The Second Xiangya Hospital of Central South University                                                                                                                   & 27               & 37      \\
5              & Department of Psychiatry,Shanghai Jiao Tong University School of Medicin                                                                                                  & 13               & 11      \\
6              & Department of Psychiatry,Shanghai Jiao Tong University School of Medicine                                                                                                 & 15               & 15      \\
7              & Sir Run Run Shaw Hospital,Zhejiang University School of Medicine                                                                                                          & 38               & 49      \\
8              & Department of Psychiatry,First Affiliated Hospital,China Medical University                                                                                               & 75               & 75      \\
9              & The First Affiliated Hospital of Jinan University                                                                                                                         & 50               & 50      \\
10             & First Hospital of Shanxi Medical University                                                                                                                               & 50               & 33      \\
11             & Department of Psychiatry,The First Affiliated Hospital of Chongqing Medical University                                                                                    & 32               & 29      \\
12             & Department of Psychiatry,The First Affiliated Hospital of Chongqing Medical University                                                                                    & 32               & 6       \\
13             & The First Affiliated Hospital of Xi an Jiaotong University,Xian Central Hospital                                                                                          & 25               & 17      \\
14             & The Second Xiangya Hospital of Central South University                                                                                                                   & 64               & 32      \\
15             & \begin{tabular}[c]{@{}l@{}}Zhongda Hospital, School of Medicine, Southeast University\end{tabular}                         & 50               & 50      \\
16             & Huaxi MR Research Center, West China Hospital of Sichuan University                                                                                                       & 31               & 31      \\
17             & Department of Psychiatry,The First Affiliated Hospital of Chongqing Medical University                                                                                    & 47               & 44      \\
18             & \begin{tabular}[c]{@{}l@{}}The First Affiliated Hospital, College of Medicine, Zhejiang University\end{tabular}                                 & 21               & 20      \\
19             & Anhui Medical University                                                                                                                                                  & 51               & 36      \\
20             & Faculty of Psychology, Southwest University                                                                                                                               & 282              & 251     \\
21             & Beijing Anding Hospital, Capital Medical University                                                                                                                       & 86               & 70      \\
22             & The Institute of Mental Health,Second Xiangya Hospital of Central South University                                                                                        & 30               & 20      \\
23             & Mental Health Center, West China Hospital, Sichuan University                                                                                                             & 32               & 30      \\
24             & First Affiliated Hospital of Kunming Medical University                                                                                                                   & 32               & 31      \\
25             & Department of Neurology,Affiliated ZhongDa Hospital of Southeast University                                                                                               & 89               & 63      \\ \bottomrule \hline
                     
\end{tabular}
\end{sidewaystable}

\begin{sidewaystable}[hbt!]
\renewcommand{\thetable}{S2}
\centering
\caption{Data acquisition parameters of selected sites.}
\label{Data acquisition parameters of selected sites}
\begin{tabular}{@{}lllllllllll@{}}
\hline
\toprule
Serial number & Scanner                       & Coil & TR (ms) & TE (ms) & Flip angle & Thickness/gap  & Slice number & Timepoints & Voxel size     & FOV     \\ \midrule \hline
1              & Siemens Tim Trio 3T           & 32   & 2000    & 30      & 90         & 4.0 mm/0.8 mm  & 30           & 210        & 3.28*3.28*4.80 & 210*210 \\
2              & Philips Achieva 3T            & 8    & 2000    & 30      & 90         & 4.0 mm/0 mm    & 37           & 200        & 1.67*1.67*4.00 & 240*240 \\
3              & Siemens 1.5 T                 & 16   & 2000    & 40      & 90         & 5.0mm/1.25mm   & 26           & 150        & 3.75*3.75*6.25 & 240*240 \\
5              & GE Signa 3T                   & 32   & 3000    & 30      & 90         & 5.0mm/0 mm     & 22           & 100        & 3.75*3.75*5.00 & 240*240 \\
6              & Siemens Tim Trio 3T           & 32   & 2000    & 30      & 70         & 4mm/0mm        & 33           & 180        & 3.59*3.59*4.00 & 230*230 \\
7              & GE discovery MR750            & 8    & 2000    & 30      & 90         & 3.2 mm/0 mm    & 37           & 184        & 2.29*2.29*3.20 & 220*220 \\
8              & GE Signa 3T                   & 8    & 2000    & 30      & 90         & 3.0 mm/0 mm    & 35           & 200        & 3.75*3.75*3.00 & 240*240 \\
9              & GE Discovery MR750 3.0T       & 8    & 2000    & 25      & 90         & 3.0 mm/1.0 mm  & 35           & 200        & 3.75*3.75*4.00 & 240*240 \\
10             & Siemens Tim Trio 3T           & 32   & 2000    & 30      & 90         & 3.0 mm/1.52 mm & 32           & 212        & 3.75*3.75*4.52 & 240*240 \\
11             & GE Signa 3T                   & 8    & 2000    & 30      & 90         & 5 mm           & 33           & 200        & 3.75*3.75*5.00 & 240*240 \\
12             & GE Signa 3T                   & 8    & 2000    & 30      & 90         & 5 mm           & 33           & 240        & 3.75*3.75*4.00 & 240*240 \\
13             & GE Excite 1.5T                & 16   & 2500    & 35      & 90         & 4 mm/0 mm      & 36           & 150        & 4.00*4.00*4.00 & 256*256 \\
14             & Siemens Tim Trio 3T           & 32   & 2500    & 25      & 90         & 3.5 mm/0 mm    & 39           & 200        & 3.75*3.75*3.50 & 240*240 \\
15             & Siemens Verio 3.0T MRI        & 12   & 2000    & 25      & 90         & 4 mm/0 mm      & 36           & 240        & 3.75*3.75*4.00 & 240*240 \\
16             & GE Signa 3T                   & 8    & 2000    & 30      & 90         & 5mm/0mm        & 30           & 200        & 3.75*3.75*5.00 & 240*240 \\
17             & GE Signa 3T                   & 8    & 2000    & 40      & 90         & 4.0 mm/0 mm    & 33           & 240        & 3.75*3.75*4.00 & 240*240 \\
18             & Philips Achieva 3.0 T scanner & 8    & 2000    & 35      & 90         & 5.0/1.0 mm     & 24           & 200        & 1.67*1.67*6.00 & 240*240 \\
19             & GE Signa 3T                   & 8    & 2000    & 22.5    & 30         & 4.0 mm/0.6 mm  & 33           & 240        & 3.44*3.44*4.60 & 220*220 \\
20             & Siemens Tim Trio 3T           & 12   & 2000    & 30      & 90         & 3.0 mm/1.0 mm  & 32           & 242        & 3.44*3.44*4.00 & 220*220 \\
21             & Siemens Tim Trio 3T           & 32   & 2000    & 30      & 90         & 3.5 mm/0.7 mm  & 33           & 240        & 3.12*3.12*4.20 & 200*200 \\
22             & Philips Gyroscan Achieva 3.0T & 32   & 2000    & 30      & 90         & 4.0 mm/0 mm    & 36           & 250        & 1.67*1.67*4.00 & 240*240 \\
23             & Philips Achieva 3.0T TX       & 8    & 2000    & 30      & 90         & 4.0 mm/0 mm    & 38           & 240        & 3.75*3.75*4.00 & 240*240 \\
24             & GE Signa 1.5T                 & 8    & 2000    & 40      & 90         & 5/1mm          & 24           & 160        & 3.75*3.75*6.00 & 240*240 \\
25             & Siemens Verio 3T              & 12   & 2000    & 25      & 90         & 4.0mm/0 mm     & 36           & 240        & 3.75*3.75*4.00 & 240*240 \\ \bottomrule \hline
\end{tabular}
\end{sidewaystable}

\clearpage

\begin{table}[hbt!]
\renewcommand{\thetable}{S3}
\centering
\caption{Demographic information of studied subjects from independent site of REST-meta-MDD.}
\label{independent site information}
\begin{tabular}{@{}lcccccc@{}}
\toprule\hline
Site  & Subjects & MDD (F)   & NC (F)    & Age (MDD)   & Age (NC)     & HAMD total scores \\ \midrule\hline
1     & 148      & 74 (43)   & 74 (42)   & 31.76±8.16  & 31.79±8.99   & 24.86±4.80        \\
2     & 60       & 30 (23)   & 30 (21)   & 43.96±12.78 & 44.6±12.38   & 23.6±3.29         \\
3     & 37       & 0         & 37 (23)   &             & 20.4±1.62    & 0                 \\
4     & 48       & 24 (13)   & 24 (12)   & 31.75±7.66  & 28.75±4.58   & 22±2.99           \\
5     & 11       & 0         & 11 (6)    &             & 33±4.95      & 0                 \\
6     & 30       & 15 (8)    & 15 (9)    & 31.33±11.98 & 28.46±10.89  & 23.13±6.25        \\
7     & 87       & 38 (24)   & 49 (30)   & 40.58±11.60 & 41.42±13.40  & 22.23±4.42        \\
8     & 150      & 75 (43)   & 75 (45)   & 32.49±13.08 & 28.68±11.09  & 22.91±8.61        \\
9     & 50       & 0         & 50 (19)   &             & 28.92±8.59   & 0                 \\
10    & 83       & 50 (35)   & 33 (13)   & 30.44±9.93  & 30.15±11.61  & 20.7±3.77         \\
11    & 37       & 8 (4)     & 29 (19)   & 29±10.92    & 32.82±9.98   & 23.37±4.37        \\
12    & 38       & 32 (16)   & 6 (1)     & 32.25±11.37 & 31.33±2.25   & 24.25±4.67        \\
13    & 42       & 25 (22)   & 17 (11)   & 35.2±9.81   & 34±10.5      & 25.24±3.89        \\
14    & 96       & 64 (38)   & 32 (17)   & 32.52±8.56  & 29.59±4.99  & 21.32±3.43       \\
15    & 100      & 50 (37)   & 50 (24)   & 37.24±15.44 & 46.48±17.64 & 27.32±5.71       \\
16    & 62       & 31 (18)   & 31 (18)   & 44.32±14.62 & 31.67±10.32 & 21.45±3.22       \\
17    & 91       & 47 (26)   & 44 (30)   & 29.51±9.69  & 20.31±2.11  & 20.78±5.47       \\
18    & 41       & 21 (15)   & 20 (12)   & 20.14±3.56  & 30.1±7.52   & 24.76±5.54       \\
19    & 87       & 51 (34)   & 36 (18)   & 29.45±9.61  & 35.86±10.26 & 22.19±8.13       \\
20    & 505      & 254 (167) & 251 (164) & 37.65±12.68 & 39.64±15.86 & 20.94±5.59       \\
21    & 155      & 85 (54)   & 70 (39)   & 32.68±12.04 & 36.12±12.63 & 14.47±8.03       \\
22    & 50       & 30 (18)   & 20 (8)    & 40.3±15.82  & 24.35±7.06  & 22.6±5.04        \\
23    & 62       & 32 (21)   & 30 (18)   & 40.71±15.35 & 32.16±13.09 & 18.56±8.40       \\
24    & 63       & 32 (19)   & 31 (20)   & 33.31±13.29 & 31.61±7.48  & 24.12±3.94       \\
25    & 152      & 89 (69)   & 63 (34)   & 63.72±10.57 & 69.63±5.86  & 5.85±4.68        \\
Total & 2285     & 1157      & 1128      & 36.96±14.65 & 35.99±15.55 & 20.53±7.60       \\ \bottomrule\hline
\end{tabular}
\begin{tablenotes}    
    \footnotesize
    \item[1] 
    F: female. HAMD: Hamilton Depression Rating Scale 
\end{tablenotes} 
\end{table}

\begin{table}[hbt!]
\renewcommand{\thetable}{S4}
\centering
\caption{The performance of MFMC on single features and multi-feature combinations.}
\label{CMFSMF}
\begin{tabular}{@{}lllll@{}}
\toprule\hline
Feature            & Accuracy (\%)  & Sensitivity (\%)& Specificity (\%)& F1 (\%)        \\ \midrule\hline
fALFF              & 53.83±2.94 & 54.77±4.30  & 52.85±6.66  & 54.72±2.77 \\
DC                 & 54.43±4.19 & 56.00±4.25  & 52.78±5.35  & 55.61±4.01 \\
VMHC               & 56.23±2.96 & 58.41±3.04  & 53.95±5.03  & 57.67±2.57 \\
fALFF+DC           & 56.98±2.38 & 54.76±3.81  & 57.66±3.87  & 55.83±2.71 \\
ReHo               & 58.97±2.63 & 58.36±3.98  & 59.60±4.73  & 58.71±2.81 \\
fALFF+VMHC         & 76.42±2.45 & 74.58±4.20  & 78.27±3.16  & 76.08±2.74 \\
DC+VMHC            & 85.61±3.14 & 82.28±3.89  & 89.02±2.98  & 85.26±3.31 \\
fALFF+VMHC+DC      & 85.99±3.67 & 83.37±4.47  & 88.76±4.92  & 85.92±3.68 \\
ReHo+DC            & 92.29±1.78 & 92.60±2.01  & 91.97±2.62  & 92.45±1.73 \\
ReHo+VMHC          & 92.78±1.28 & 90.91±3.00  & 94.68±2.04  & 92.70±1.40 \\
ReHo+DC+fALFF      & 93.54±1.85 & 93.37±3.06  & 93.71±1.78  & 93.56±1.90 \\
ReHo+fALFF         & 94.58±2.00 & 94.47±3.00  & 94.70±2.73  & 94.72±1.98 \\
ReHo+VMHC+fALFF    & 96.23±1.53 & 95.31±2.33  & 97.19±2.09  & 96.28±1.51 \\
ReHo+VMHC+DC       & 96.50±0.98 & 95.19±2.03  & 97.81±1.20  & 96.45±1.04 \\
ReHo+DC+fALFF+VMHC & 96.88±0.85 & 96.40±1.82  & 97.38±1.77  & 96.87±0.86 \\ \bottomrule\hline
\end{tabular}
\end{table}

\begin{table}[hbt!]
\renewcommand{\thetable}{S5}
\centering
\caption{The performance of the kNN on single features and multi-level combinations.}
\label{kNN}
\begin{tabular}{@{}lllll@{}}
\toprule\hline
Feature            & Accuracy (\%)   & Sensitivity (\%) & Specificity (\%) & F1 (\%)         \\ \midrule\hline
fALFF              & 55.36±2.71 & 57.12±4.86  & 53.52±4.55  & 56.57±3.24 \\
fALFF+DC           & 55.80±3.61 & 53.13±3.91  & 58.53±5.37  & 54.85±3.92 \\
DC                 & 55.97±3.53 & 55.97±5.84  & 55.37±7.50  & 55.37±7.50 \\
ReHo               & 56.29±3.89 & 55.30±7.03  & 57.28±4.68  & 55.73±4.91 \\
VMHC               & 56.67±2.61 & 52.84±4.89  & 60.68±5.81  & 55.37±3.25 \\
fALFF+VMHC         & 66.63±3.59 & 56.78±5.64  & 76.64±4.56  & 63.06±4.51 \\
DC+VMHC            & 70.78±2.78 & 53.35±3.63  & 88.69±4.02  & 64.88±3.46 \\
fALFF+VMHC+DC      & 72.92±3.04 & 62.06±4.67  & 84.38±4.20  & 70.10±3.79 \\
ReHo+DC            & 74.18±3.71 & 59.23±6.08  & 89.73±2.98  & 69.92±5.05 \\
ReHo+fALFF         & 74.78±2.43 & 60.78±4.16  & 89.62±2.64  & 71.21±3.28 \\
ReHo+DC+fALFF      & 75.88±1.99 & 61.34±3.32  & 90.63±2.41  & 71.89±2.65 \\
ReHo+VMHC          & 76.64±2.47 & 61.90±4.48  & 91.70±2.01  & 72.73±3.54 \\
ReHo+VMHC+fALFF    & 79.38±2.82 & 68.37±5.04  & 90.92±2.57  & 77.16±3.51 \\
ReHo+VMHC+DC       & 81.02±3.58 & 67.98±5.88  & 94.09±2.14  & 78.08±4.60 \\
ReHo+DC+fALFF+VMHC & 81.89±2.83 & 70.68±4.75  & 93.10±2.77  & 79.55±3.46 \\ \bottomrule\hline
\end{tabular}
\end{table}

\begin{table}[hbt!]
\renewcommand{\thetable}{S6}
\centering
\caption{The performance of the QDA on single features and multi-level combinations.}
\label{QDA}
\begin{tabular}{@{}lllll@{}}
\toprule\hline
Feature            & Accuracy (\%)  & Sensitivity (\%) & Specificity (\%) & F1 (\%)        \\ \midrule\hline
fALFF              & 54.00±3.31 & 29.59±8.14  & 79.44±6.62  & 39.07±7.65 \\
DC                 & 54.43±3.32 & 59.76±12.82 & 48.91±9.48  & 56.47±7.68 \\
fALFF+DC           & 55.14±2.62 & 59.73±6.19  & 50.45±6.27  & 57.30±3.52 \\
VMHC               & 57.44±2.89 & 61.10±4.16  & 53.64±4.80  & 59.41±2.91 \\
ReHo               & 58.81±3.20 & 60.11±7.64  & 57.50±7.04  & 59.18±4.32 \\
fALFF+VMHC         & 76.15±2.37 & 63.83±5.55  & 88.64±2.49  & 72.82±3.58 \\
DC+VMHC            & 81.34±3.54 & 71.60±6.44  & 91.35±2.22  & 79.40±4.49 \\
fALFF+VMHC+DC      & 83.26±3.19 & 77.30±5.20  & 89.55±3.06  & 82.50±2.86 \\
ReHo+DC            & 89.49±2.50 & 92.70±3.69  & 86.16±4.12  & 89.99±2.36 \\
ReHo+VMHC          & 91.30±1.29 & 85.17±2.22  & 97.56±1.30  & 90.81±1.43 \\
ReHo+VMHC+DC       & 91.41±1.29 & 83.60±2.98  & 99.23±0.86  & 90.67±1.55 \\
ReHo+DC+fALFF      & 91.58±2.35 & 96.31±2.02  & 86.78±3.82  & 92.03±2.14 \\
ReHo+VMHC+fALFF    & 91.85±1.55 & 85.16±2.45  & 98.88±1.74  & 91.45±1.65 \\
ReHo+fALFF         & 92.17±2.48 & 96.28±1.45  & 87.81±3.96  & 92.71±2.23 \\
ReHo+DC+fALFF+VMHC & 93.00±1.97 & 95.58±2.28  & 90.35±3.28  & 93.27±1.86 \\ \bottomrule\hline
\end{tabular}
\end{table}

\begin{table}[hbt!]
\renewcommand{\thetable}{S7}
\centering
\caption{The performance of the XGBoost on single features and multi-level combinations.}
\label{XGBoost}
\begin{tabular}{@{}lllll@{}}
\toprule\hline
Feature            & Accuracy (\%)  & Sensitivity (\%) & Specificity (\%) & F1 (\%)         \\ \midrule\hline
fALFF              & 55.03±2.11 & 56.05±3.86  & 53.98±4.52  & 55.95±2.42 \\
DC                 & 55.96±3.53 & 57.51±5.28  & 44.18±6.59  & 60.70±2.82 \\
fALFF+DC           & 56.84±3.38 & 55.72±3.52  & 57.98±3.75  & 56.63±3.72 \\
ReHo               & 58.15±2.61 & 57.27±3.62  & 59.03±4.97  & 57.79±2.60 \\
VMHC               & 58.53±1.94 & 59.59±3.31  & 57.43±5.23  & 59.44±1.82 \\
fALFF+VMHC         & 68.16±4.56 & 63.93±6.92  & 72.43±4.41  & 66.80±5.24 \\
DC+VMHC            & 72.64±3.07 & 66.19±3.23  & 79.26±4.87  & 71.04±3.11 \\
fALFF+VMHC+DC      & 74.78±2.66 & 69.83±5.10  & 80.00±4.29  & 73.90±3.12 \\
ReHo+DC            & 75.88±2.34 & 70.18±3.39  & 81.80±2.77  & 74.77±2.57 \\
ReHo+fALFF         & 77.40±3.35 & 73.53±5.80  & 81.51±3.77  & 76.92±3.90 \\
ReHo+VMHC          & 78.67±2.69 & 72.61±4.26  & 84.84±3.91  & 77.44±3.04 \\
ReHo+DC+fALFF      & 80.47±2.26 & 74.27±3.64  & 86.77±2.17  & 79.27±2.58 \\
ReHo+VMHC+fALFF    & 84.57±3.44 & 80.46±4.12  & 88.89±4.58  & 84.23±3.49 \\
ReHo+VMHC+DC       & 85.28±1.97 & 79.24±4.30  & 91.35±3.31  & 84.31±2.38 \\
ReHo+DC+fALFF+VMHC & 85.72±2.77 & 80.10±2.74  & 91.25±2.67  & 84.89±2.81 \\ \bottomrule\hline
\end{tabular}
\end{table}

\begin{table}[hbt!]
\renewcommand{\thetable}{S8}
\centering
\caption{The performance of the stacking method (meta classifier kNN) on single features and multi-feature combinations.}
\label{CMFSMF_meta_knn}
\begin{tabular}{@{}lllll@{}}
\toprule\hline
Feature            & Accuracy (\%)  & Sensitivity (\%) & Specificity (\%) & F1 (\%)        \\ \midrule\hline
DC                 & 55.83±3.20 & 53.85±3.29  & 57.84±4.63  & 55.17±2.98 \\
ReHo               & 56.56±2.71 & 56.67±3.91  & 56.44±4.65  & 56.88±2.88 \\
fALFF              & 56.80±3.10 & 58.65±5.30  & 54.90±2.74  & 57.82±3.96 \\
VMHC               & 57.56±4.05 & 61.54±6.38  & 53.47±3.34  & 59.53±5.20 \\
fALFF+DC           & 57.63±4.33 & 59.27±6.58  & 58.45±4.57  & 57.55±4.25 \\
fALFF+VMHC         & 66.59±3.06 & 53.03±4.02  & 80.51±3.65  & 61.61±3.81 \\
DC+VMHC            & 71.94±3.14 & 58.59±4.60  & 85.62±2.54  & 67.82±4.17 \\
fALFF+VMHC+DC      & 73.01±2.72 & 58.21±5.70  & 88.18±2.76  & 68.45±2.19 \\
ReHo+VMHC          & 77.63±2.74 & 63.11±4.91  & 92.52±2.01  & 73.98±3.79 \\
ReHo+DC            & 78.06±1.99 & 66.37±4.25  & 90.05±2.64  & 75.33±2.46 \\
ReHo+fALFF         & 78.99±3.80 & 66.48±6.96  & 91.83±2.91  & 76.03±5.35 \\
ReHo+DC+fALFF      & 79.33±2.80 & 66.48±5.79  & 92.51±1.83  & 76.37±4.01 \\
ReHo+VMHC+fALFF    & 79.67±2.22 & 65.61±4.59  & 94.09±2.62  & 76.49±3.09 \\
ReHo+VMHC+DC       & 81.24±2.67 & 68.59±4.69  & 94.24±1.91  & 78.67±3.51 \\
ReHo+DC+fALFF+VMHC & 83.06±2.55 & 72.83±4.80  & 93.41±2.23  & 81.21±0.47 \\ \bottomrule\hline
\end{tabular}
\end{table}

\begin{table}[hbt!]
\renewcommand{\thetable}{S9}
\centering
\caption{The performance of the stacking method (meta classifier QDA) on single features and multi-feature combinations.}
\label{CMFSMF_meta_QDA}
\begin{tabular}{@{}lllll@{}}
\toprule\hline
Feature            & Accuracy (\%)  & Sensitivity (\%)& Specificity (\%)& F1 (\%)        \\ \midrule\hline
DC                 & 54.57±2.77 & 63.69±5.12  & 45.22±6.91  & 58.61±2.79 \\
fALFF              & 55.69±2.11 & 42.65±2.43  & 69.06±3.60  & 49.35±2.28 \\
fALFF+DC           & 55.98±3.28 & 47.64±5.13  & 61.46±5.38  & 51.36±3.82 \\
VMHC               & 58.54±4.87 & 66.35±4.60  & 50.50±6.35  & 61.88±4.36 \\
ReHo               & 59.51±2.60 & 53.85±3.92  & 65.35±2.72  & 57.44±3.26 \\
fALFF+VMHC         & 77.53±3.82 & 79.83±4.77  & 75.16±3.90  & 78.22±3.80 \\
DC+VMHC            & 81.28±3.09 & 84.82±2.69  & 77.64±4.38  & 82.12±2.83 \\
fALFF+VMHC+DC      & 85.92±3.07 & 76.19±5.79  & 96.04±1.25  & 84.66±4.17 \\
ReHo+VMHC+DC       & 91.03±1.41 & 96.87±1.83  & 85.04±2.72  & 91.63±1.28 \\
ReHo+DC            & 91.22±2.46 & 93.27±2.98  & 89.11±3.72  & 91.51±2.37 \\
ReHo+DC+fALFF      & 91.78±2.44 & 96.45±1.87  & 87.01±4.00  & 92.26±2.23 \\
ReHo+VMHC          & 92.68±1.34 & 93.27±3.08  & 92.08±2.94  & 92.82±1.86 \\
ReHo+fALFF         & 93.29±1.87 & 95.20±2.23  & 91.34±2.65  & 93.49±1.81 \\
ReHo+VMHC+fALFF    & 93.43±1.15 & 97.12±1.36  & 89.66±2.29  & 93.75±1.07 \\
ReHo+DC+fALFF+VMHC & 94.16±1.60 & 89.52±3.00  & 98.92±0.53  & 93.92±1.76 \\ \bottomrule\hline
\end{tabular}
\end{table}

\clearpage

\begin{table}[hbt!]
\renewcommand{\thetable}{S10}
\centering
\caption{The performance of MFMC in combined sites, in which the number of subjects are close to that under 5-fold cross-validation. $\Delta_\text{accuracy}$ measures the difference between the accuracy in combined sites and the accuracy under 5-fold cross-validation (96.88\% in Table \ref{CMFSMF}).}
\label{MFMC combined sites}
\begin{tabular}{@{}lccccccc@{}}
\toprule\hline
Testset                       & Accuracy (\%)& Sensitivity (\%)& Specificity (\%)& F1 (\%)   & $\Delta_\text{accuracy}$ (\%)& NC  & MDD \\ \midrule\hline
Site2, 8, 15, 16, 18 \& 23        & 95.85    & 96.33       & 95.37       & 95.89 & -1.03      & 236 & 239 \\
Site2, 4, 6, 8, 16, 22 \& 24  & 94.60     & 94.51       & 94.69       & 94.71 & -2.28      & 226 & 237 \\
Site3, 8, 11, 12 \& 14        & 94.41    & 97.21       & 91.62       & 94.57 & -2.47      & 179 & 179 \\
Site2, 7, 8, 18 \& 21         & 93.31    & 93.98       & 92.62       & 93.41 & -3.57      & 244 & 249 \\
Site5, 8, 9, 10, 12, 13 \& 19 & 93.28    & 93.56       & 92.98       & 93.36 & -3.6       & 228 & 233 \\
Site20                        & 90.93    & 97.24       & 83.27       & 90.98 & -5.95       & 251 & 254 \\
Site1, 9, 13, 19 \& 25        & 87.89    & 82.01       & 93.75       & 87.11 & -8.99      & 240 & 239 \\
Site1, 2, 7, 18 \& 21         & 87.17    & 79.44       & 95.06       & 86.21 & -9.71      & 243 & 248 \\ \bottomrule\hline
\end{tabular}
\end{table}

\begin{table}[hbt!]
\renewcommand{\thetable}{S11}
\centering
\caption{The performance of kNN in combined sites. $\Delta_\text{accuracy}$ measures the difference between the accuracy in combined sites and the accuracy under 5-fold cross-validation (81.99\% in Table \ref{kNN}).}
\label{kNN combined sites}
\begin{tabular}{@{}lccccccc@{}}
\toprule\hline
Testset                       & Accuracy (\%)& Sensitivity (\%)& Specificity (\%)& F1 (\%)   & $\Delta_\text{accuracy}$ (\%)& NC  & MDD \\ \midrule\hline
Site2, 8, 15, 16,18,23        & 71.43    & 45.41       & 97.69       & 61.49 & -10.46      & 236 & 239 \\
Site2, 4, 6, 8, 16, 22 \& 24  & 72.14    & 48.95       & 96.46       & 64.27 & -9.75       & 226 & 237 \\
Site3, 8, 11, 12 \& 14        & 74.02    & 53.63       & 94.41       & 67.37 & -7.87       & 179 & 179 \\
Site2, 7, 8, 18 \& 21         & 69.17    & 45.38       & 93.44       & 59.79 & -12.72      & 244 & 249 \\
Site5, 8, 9, 10, 12, 13 \& 19 & 74.62    & 58.80       & 90.79       & 70.08 & -7.27       & 228 & 233 \\
Site20                        & 66.53    & 59.06       & 74.10       & 63.97 & -15.36      & 251 & 254 \\
Site1, 9, 13, 19 \& 25        & 70.98    & 50.63       & 91.25       & 63.52 & -10.91      & 240 & 239 \\
Site1, 2, 7, 18 \& 21         & 66.80    & 40.32       & 93.83       & 55.10 & -15.09      & 243 & 248 \\ \bottomrule\hline
\end{tabular}
\end{table}

\begin{table}[hbt!]
\renewcommand{\thetable}{S12}
\centering
\caption{The performance of QDA in combined sites. $\Delta_\text{accuracy}$ measures the difference between the accuracy in combined sites and the accuracy under 5-fold cross-validation (93.00\% in Table \ref{QDA}).}
\label{QDA combined sites}
\begin{tabular}{@{}lccccccc@{}}
\toprule\hline
Testset                       & Accuracy (\%)& Sensitivity (\%)& Specificity (\%)& F1 (\%)   & $\Delta_\text{accuracy}$ (\%)& NC  & MDD \\ \midrule\hline
Site2, 8, 15, 16,18,23        & 71.66    & 99.54       & 43.52       & 77.92 & -21.34      & 236 & 239 \\
Site2, 4, 6, 8, 16, 22 \& 24  & 74.30     & 99.16       & 48.23       & 79.80  & -18.70       & 226 & 237 \\
Site3, 8, 11, 12 \& 14        & 73.18    & 99.44       & 46.93       & 78.76 & -19.82      & 179 & 179 \\
Site2, 7, 8, 18 \& 21         & 78.09    & 99.6        & 56.15       & 82.12 & -14.91      & 244 & 249 \\
Site5, 8, 9, 10, 12, 13 \& 19 & 69.63    & 99.57       & 39.04       & 76.82 & -23.37      & 228 & 233 \\
Site20                        & 79.01    & 1           & 57.77       & 82.74 & -13.99      & 251 & 254 \\
Site1, 9, 13, 19 \& 25        & 73.07    & 91.21       & 55.00          & 77.17 & -19.93      & 240 & 239 \\
Site1, 2, 7, 18 \& 21         & 78.62    & 88.31       & 68.72       & 80.66 & -14.38      & 243 & 248 \\ \bottomrule\hline
\end{tabular}
\end{table}

\begin{table}[hbt!]
\renewcommand{\thetable}{S13}
\centering
\caption{The performance of XGBoost in combined sites. $\Delta_\text{accuracy}$ measures the difference between the accuracy in combined sites and the accuracy under 5-fold cross-validation (85.72\% in Table \ref{XGBoost}).}
\label{XGBoost combined sites}
\begin{tabular}{@{}lccccccc@{}}
\toprule\hline
Testset                       & Accuracy (\%)& Sensitivity (\%)& Specificity (\%)& F1 (\%)   & $\Delta_\text{accuracy}$ (\%)& NC  & MDD \\ \midrule\hline
Site2, 8, 15, 16, 18, \& 23   & 79.03    & 70.64       & 87.50        & 77.19 & -6.69       & 236 & 239 \\
Site2, 4, 6, 8, 16, 22 \& 24  & 83.80     & 76.37       & 91.59       & 82.84 & -1.92       & 226 & 237 \\
Site3, 8, 11, 12 \& 14        & 74.58    & 77.09       & 72.07       & 81.20  & -11.14      & 179 & 179 \\
Site2, 7, 8, 18 \& 21         & 78.56    & 63.05       & 88.93       & 72.52 & -7.16       & 244 & 249 \\
Site5, 8, 9, 10, 12, 13 \& 19 & 80.48    & 78.11       & 82.89       & 80.18 & -5.24       & 228 & 233 \\
Site20                        & 74.46    & 83.86       & 64.94       & 76.76 & -11.26      & 251 & 254 \\
Site1, 9, 13, 19 \& 25        & 74.74    & 60.67       & 88.75       & 70.56 & -10.98      & 240 & 239 \\
Site1, 2, 7, 18 \& 21         & 68.23    & 48.79       & 88.07       & 60.80  & -17.49      & 243 & 248 \\ \bottomrule\hline
\end{tabular}
\end{table}

\clearpage

\begin{table}[hbt!]
\renewcommand{\thetable}{S14}
\centering
\caption{The performance of MFMC on low-dimensional features.}
\label{The results of MFMC on low-dimensional features}
\begin{tabular}{@{}lllll@{}}
\toprule\hline
Features                                     & Accuracy (\%)& Sensitivity (\%)& Specificity (\%)& F1 (\%)    \\ \midrule\hline
Top 20 features of SHAP values               & 87.03   & 83.69      & 90.47      & 86.71 \\
Top 13 features with significant differences & 84.52   & 81.10       & 88.01      & 84.14 \\
Top 13 features of SHAP values               & 83.86   & 79.81      & 88.03      & 83.6 \\
Top 10 features with significant differences & 83.42   & 78.51      & 88.47      & 82.71 \\
Top 10 features of SHAP values               & 78.78   & 74.08      & 83.59      & 77.95 \\
Top 4 features with significant differences  & 61.21   & 60.04      & 62.42      & 60.99 \\
Top 4 features of SHAP values                & 60.12   & 59.16      & 61.09      & 59.99 \\ \bottomrule\hline
\end{tabular}
\end{table}

\clearpage
\end{document}